\documentclass[10pt,twocolumn,letterpaper]{article}

\usepackage{cvpr}              % To produce the CAMERA-READY version
\usepackage{graphicx}
\usepackage{amsmath}
\usepackage{amssymb}
\usepackage{booktabs}
\usepackage{adjustbox}
\usepackage{microtype}
\usepackage{booktabs}
\usepackage{ulem}
\usepackage{cuted}
\usepackage{multirow}
\AfterEndEnvironment{strip}{\leavevmode}

\usepackage[pagebackref,breaklinks,colorlinks]{hyperref}

\newcommand*\samethanks[1][\value{footnote}]{\footnotemark[#1]}

\usepackage[capitalize]{cleveref}
\crefname{section}{Sec.}{Secs.}
\Crefname{section}{Section}{Sections}
\Crefname{table}{Table}{Tables}
\crefname{table}{Tab.}{Tabs.}
\newcommand{\myparagraph}[1]{\vspace{2pt}\noindent{\bf #1}}

\def\cvprPaperID{****}
\def\confName{}
\def\confYear{2023}

\begin{document}

\title{LISA: Localized Image Stylization with Audio \\ via Implicit Neural Representation}

\author{
Seung Hyun Lee$^1$\thanks{Equal contribution} \quad Chanyoung Kim$^1$\samethanks \quad Wonmin Byeon$^2$ \quad Sang Ho Yoon$^3$\\
 Jinkyu Kim$^{4}$\thanks{Corresponding author} \quad Sangpil Kim$^{1}$\samethanks \\ 
  $^1$Department of Artificial Intelligence and $^4$CSE, Korea University\\
  $^2$NVIDIA Research, NVIDIA Corporation\\
  $^3$Graduate School of Culture Technology, KAIST
}
\maketitle

\begin{strip}\centering
% \vspace{-4.em}
\vspace{-6em}
\includegraphics[width=\textwidth]{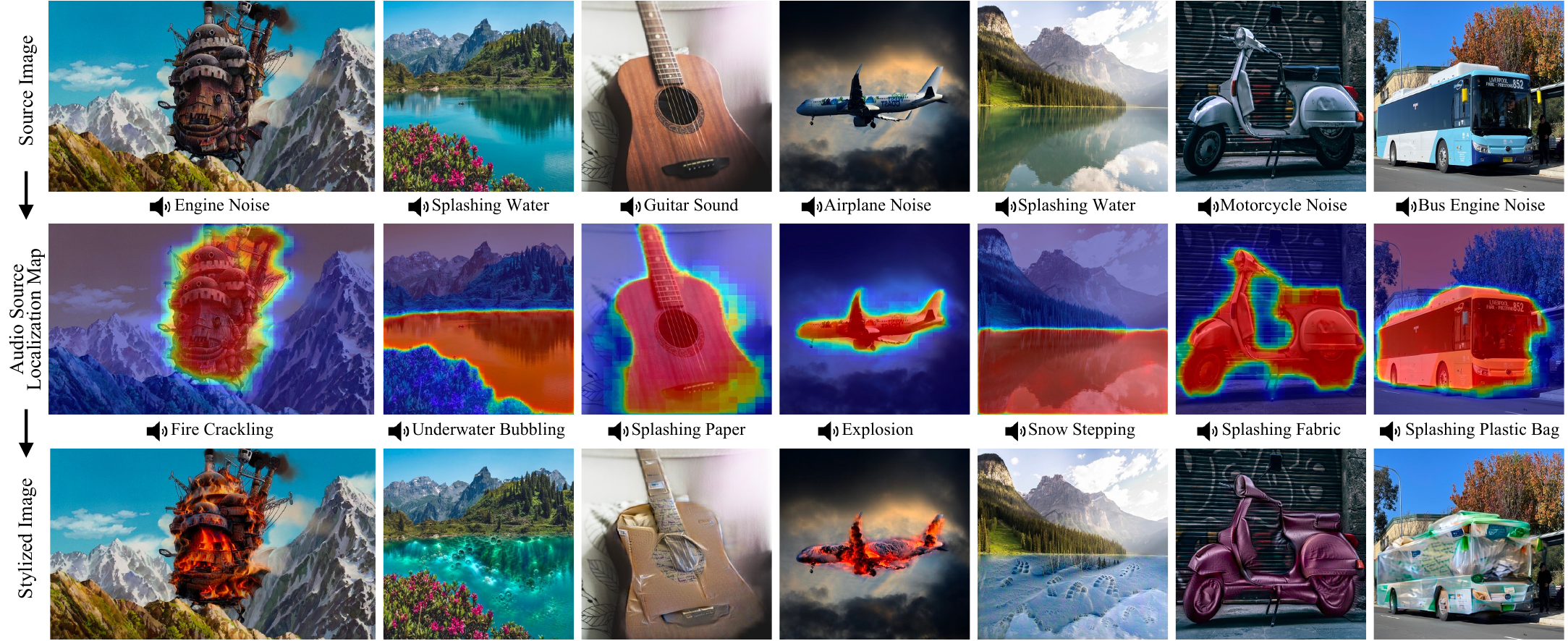}
\captionof{figure}
{
Examples from our localized image stylization based on audio inputs. Our model first localizes image regions corresponding to an audio input (e.g., given the sound of splashing water, our model localizes water from a source image). Conditioned on such a localization map, our model further stylizes the source image driven by another audio source (e.g., given an underwater bubbling sound, our model stylizes water as bubbling water).
}
\label{fig:fig0}
\vspace{-1em}
\end{strip}

\begin{abstract}
    \vspace{-1em}
    We present a novel framework, Localized Image Stylization with Audio~\textbf{(LISA)} which performs audio-driven localized image stylization. 
    Sound often provides information about the specific context of the scene and is closely related to a certain part of the scene or object. However, existing image stylization works have focused on stylizing the entire image using an image or text input. Stylizing a particular part of the image based on audio input is natural but challenging.
    In this work, we propose a framework that a user provides an audio input to localize the sound source in the input image and another for locally stylizing the target object or scene. LISA first produces a delicate localization map with an audio-visual localization network by leveraging CLIP embedding space. We then utilize implicit neural representation~(INR) along with the predicted localization map to stylize the target object or scene based on sound information. The proposed INR can manipulate the localized pixel values to be semantically consistent with the provided audio input.
    Through a series of experiments, we show that the proposed framework outperforms the other audio-guided stylization methods. Moreover, LISA constructs concise localization maps and naturally manipulates the target object or scene in accordance with the given audio input. 
    % \vspace{-1.5em}
\end{abstract}

%%%%%%%%% BODY TEXT
\section{Introduction}
Audio-guided image stylization transfers the style from sound to an image without changing the content of the original image. While recent audio-guided image stylization works show promising results by learning the audio-visual relationship~\cite{lee2022sound, li2021learning, lee2022robust}, they modify the entire image's style, which makes the output unrealistic if the image content does not match with the target style. Also, these approaches are unsuitable for the image editing application since users cannot select the desired content to stylize.

To alleviate these issues, early approaches leverage a segmentation map to localize the stylizing area after changing the style of the entire image~\cite{kurzman2019class, castillo2017zorn, stahl2019style, virtusio2018interactive}.
As these segmentation maps are often obtained by a pre-trained supervised semantic segmentation network, the segmentation area is limited by the pre-defined object categories. 
% Also, the boundary is not smooth because those works adhere to cut and paste the segmentation area of stylized image.
Moreover, previous approaches directly copy the background part from the source image to the stylized image using a binary mask, thus the final results look unnatural with rough object boundaries. 

As sound is typically associated with a specific part of the scene or an object, this information can be used to localize the area. Recent studies~\cite{chen2021localizing,10.1007/978-3-031-19836-6_13} show that the model can learn to localize sound source objects by aligning audio and visual features without any ground-truth data. However, these approaches use a class-specific activation map or a heatmap learned by audio-visual features to locate the sound source object. Therefore, the localization is not pixel-level accurate which affects the quality of stylization. 
In this work, we introduce a pixel-level localization based on input sound which is used for stylizing the image locally. 

 Furthermore, many image stylization approaches~\cite{johnson2016perceptual, 7780634,luan2017deep,huang2017arbitrary,kwon2022clipstyler} use convolution neural network~(CNN) architecture as a backbone to stylize the image. These methods often produce boundary artifacts, as shown in Fig.~\ref{fig:fig7_1}, since they do not support direct pixel-wise mapping. 
 To reduce the artifacts, we design stylizing network as Implicit Neural Representation~(INR)~\cite{tancik2020fourier} to generate per-pixel values based on audio input given corresponding pixel coordinates.

The proposed framework consists of two stages.
First, we design an audio-visual localizer to localize sounding objects or scenes in an image. 
Specifically, we train the audio-visual localizer to predict segmentation maps with pseudo ground-truth by zero-shot text-based segmentation networks in a weakly supervised manner.
Second, the localized object or scene is stylized based on the semantics of the input sound via INR which maps each RGB pixel value from the coordinate that corresponds to the semantic meaning of the input sound. 
Finally, given the localization map as the probability mask predicted by the audio-visual localizer, the style of the source image is manipulated to match the semantics of the audio by leveraging the multi-modal shared embedding space.

In our extensive experiments, our method outperforms other localized~\cite{kurzman2019class}, text-guided~\cite{kwon2022clipstyler} and sound-guided~\cite{lee2022sound, li2021learning} state-of-the-art methods. As shown in Fig.~\ref{fig:fig0}, 
given the \textit{Engine} sound input and source image named \textit{``Howl's Moving Castle''}, the castle is localized. Then a style of burning castle is produced with the fire crackling sound with a subsequent audio input.  We demonstrate various examples in the supplemental material and project website. In this study, our main contributions are listed as follows:

\begin{itemize}
    \item We propose a novel audio-guided image localized stylization. In particular, our method is able to perform partial stylization without manually producing a segmentation mask by using sound source localization. \vspace{-.5em}
    \item We carefully design implicit neural representation for style transfer to achieve naturalness. We demonstrate that this structure is effective in removing boundary artifacts. With INR, our method can stylize the source image at any arbitrary resolution. \vspace{-.5em}
    \item We achieve state-of-the-art performance compared with the existing sound source visual localization methods. Using a pseudo-ground-truth segmentation mask from images and text joint representation, we improve the quality of the audio-visual source localization that generates concise object or scene boundaries. \vspace{-.5em}
\end{itemize}
\begin{figure*}[t!]
    % \vspace{-1.59cm}
    \begin{center}
        \includegraphics[width=\textwidth]{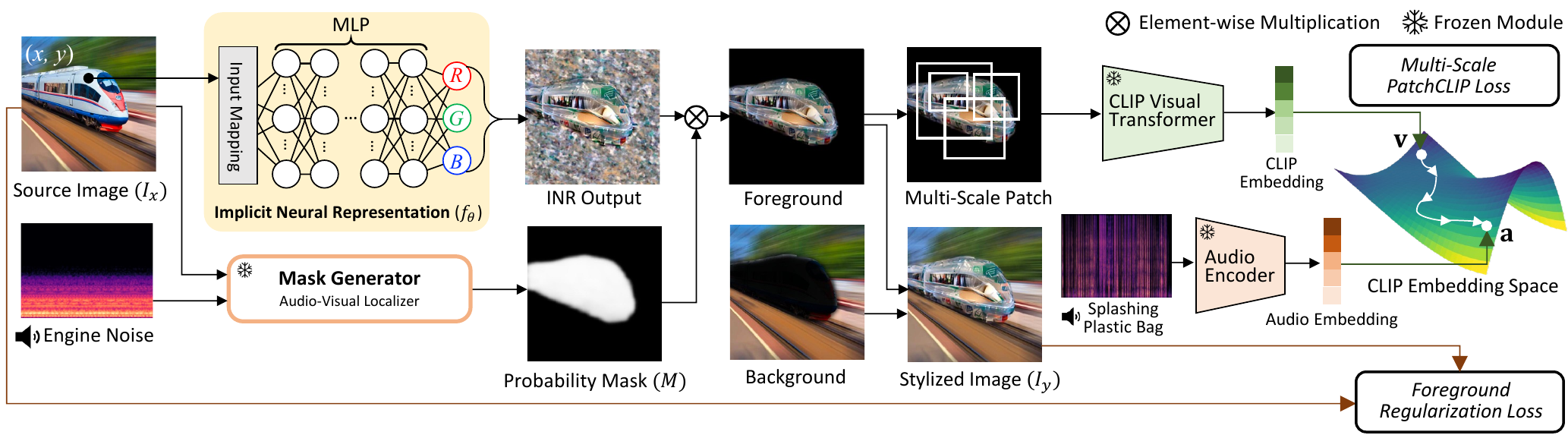}
    \end{center}
    \vspace{-1em}
    \caption{An overview of our proposed method called Localized Image Stylization with Audio (LISA). Our model consists of two main parts: (i) Audio-Visual Localizer, which outputs a pixel-level localization mask conditioned on an audio input (e.g., given a sound of engine noise input, our model localizes a train from the source image, producing a probability mask) and (ii) Audio-Guided INR Stylizer, which outputs stylized images by taking pixel locations as input and producing RGB pixel values as output. Conditioned on a new user-provided sound input (e.g., splashing plastic bag), our model is optimized with multi-scale PatchCLIP loss to generate an audio-guided “locally” stylized image. We also use Foreground Regularization Loss to make the stylized image and a source image perceptually look similar.}
    \label{fig:overview}
    \vspace{-1em}
\end{figure*}

\section{Related Work}

\myparagraph{Audio-Driven Image Stylization.} 
Audio-driven image manipulation approaches have demonstrated the usefulness of sound for manipulating the style and appearance of objects in an image. A few works show promising visual synthesis results with sound~\cite{son2017lip, deng2021unsupervised}. 
A typical task to generate from a speech sound is a neural talking head~\cite{chen2018lip, zhou2019talking, zakharov2019few, zhou2020makelttalk}, which aims to synchronize sound and head with various priors related to the face, such as keypoints and head pose. Lee~\textit{et al.} manipulates the image to be suitable with the semantics of the sound, such as~\textit{fire crackling} by leveraging pre-trained CLIP-based~\cite{radford2021learning} audio embedding space in StyleGAN latent space~\cite{lee2022sound}. Li~\textit{et al.} shows successful audio-guided image stylization~\cite{li2021learning} in an unsupervised manner. However, those works are not able to perform localized image editing. Instead, we can edit the images with the sound semantics and sound source localization using the audio-visual localizer.

\myparagraph{Localized Image Style Transfer.} 
Localized image style transfer is defined as stylizing specific parts of the source image instead of the whole image. Gatys~\textit{et al.}~\cite{gatys2017controlling} utilize the predefined binary mask for spatial control of stylization. Alegre~\textit{et al.}~\cite{alegre2020selfieart} proposed a framework to stylize the source image using facial semantic segmentation. Castillo~\textit{et al.}~\cite{castillo2017zorn} and Virtusio~\textit{et al.}~\cite{virtusio2018interactive} present a method that smoothly merges the extracted stylized object with the background.  
% also proposed a user-specified object-oriented style transfer with the pre-trained segmentation network for artistic results. 
Xia~\textit{et al.}~\cite{xia2021real} also propose the mask enhancement network to smooth the edge of the mask. 
Despite the promising results, the methods above are highly dependent on the class-based pre-trained models, which do not consider how to obtain the mask in the real world.
Text2LIVE~\cite{bar2022text2live} is the first to adopt the mask-free approach for localized image editing by text prompt. However, it is difficult for users to find appropriate text queries about localization in real-world scenarios. Using an audio-visual localizer, we can automatically find audio-related objects in the image. 

\myparagraph{Audio-Visual Sound Source Localization.} 
A few works demonstrate that audio-visual localization can localize the region of the image where the sound comes from without ground truth~\cite{tian2018audio,10.1007/978-3-031-19836-6_13,chen2021localizing}. LVS~\cite{chen2021localizing} proposes a mechanism that mines hard negative samples to formulate contrastive learning with a differentiable threshold. EZ-VSL~\cite{10.1007/978-3-031-19836-6_13} introduces multiple instances contrastive learning to focus on the most aligned regions in audio-visual representation. However, those works have a limitation in that they cannot create a delicate audio-visual source localization map because they learn to produce a localization map without any supervision. Zhou~\textit{et al.}~\cite{zhou2022audio} proposed the pixel-level segmentation of the sounding object in a supervised manner. To overcome the class-limited problem, we propose a weakly supervised audio-visual source localization by leveraging a powerful knowledge of zero-shot text-based  segmentation~\cite{luddecke2022image}.

\myparagraph{Implicit Neural Representations~(INRs).} Neural implicit networks are defined as fields parameterized by multi-layer perceptrons and have the advantage of being able to encode a continuous signal at an arbitrary resolution~\cite{xie2022neural}. INRs recently have shown that neural fields can be represented successfully for various domains with coordinates as input~\cite{genova2019learning,genova2020local,park2019deepsdf, sitzmann2020implicit}.
This approach can capture high-frequency signals, which conventional convolutional neural networks cannot. In addition, implicit neural representations have the advantage of fast optimization speed and a small number of parameters.
Following the powerful novel view synthesis capabilities of neural radiance fields, many style transfer studies focus on producing a stylized novel view~\cite{mu20223d,chiang2022stylizing,huang2022stylizednerf,hollein2022stylemesh}. Inspired by these works, we propose implicit neural representations for audio-guided localized stylization.

\section{Method}
In this paper, we propose a novel audio-guided localized image stylization method. As shown in Fig.~\ref{fig:overview},  our model has two main components: (i) Audio-Visual Localizer (Sec.~\ref{sec:zero}) and (ii) Audio-guided INR Stylizer (Sec.~\ref{sec:local}). Our {\it Audio-visual Localizer} predicts a pixel-level localization mask given a sound input (e.g., given a sound of ``engine noise,'' our model localizes a train from a source image). Conditioned on this localization map, our {\it Audio-guided INR Stylizer} outputs a ``locally'' stylized image according to the semantics of the new user-provided audio sources (e.g., given a sound of ``a splashing plastic bag,'' our model stylizes a train from a source image accordingly).

\subsection{Audio-Visual Localizer}
\label{sec:zero}
\begin{figure}[t!]
  \centering
  \includegraphics[width = \linewidth]{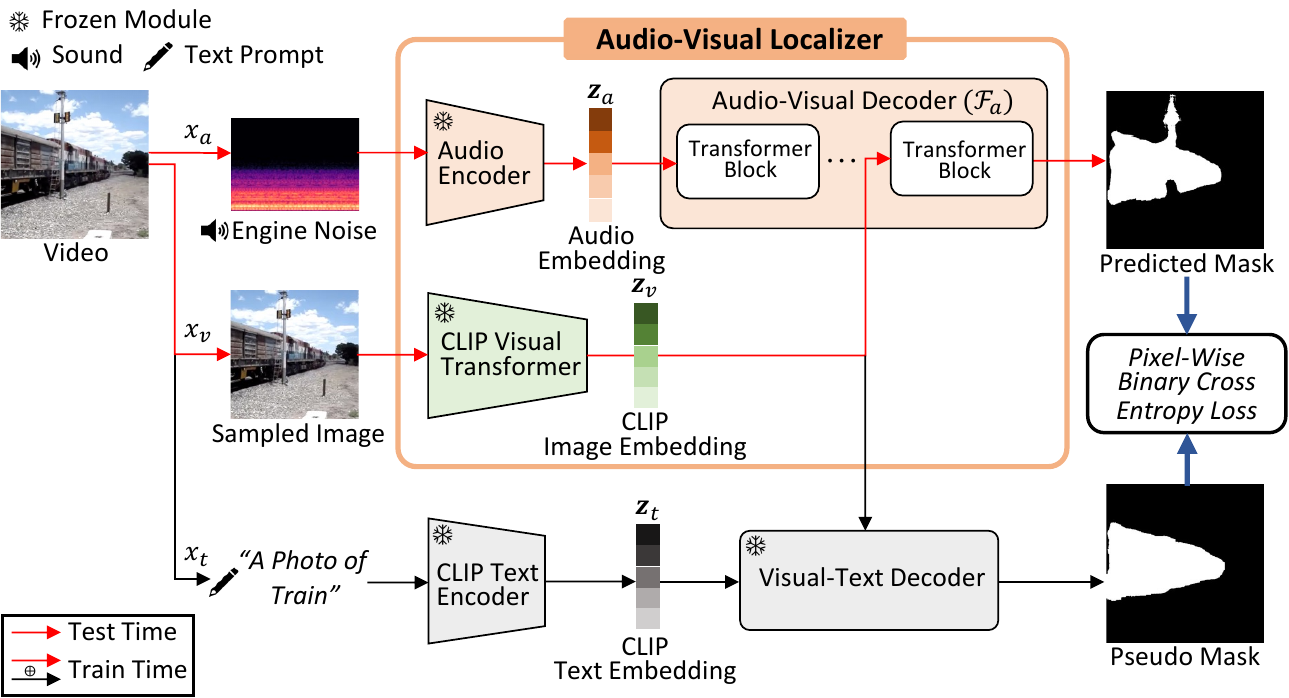}
  \caption{An overview of our Audio-visual Localizer, which identifies an image region corresponding to the sound input (e.g., localizing a train based on a sound of engine noise), producing a probability mask as an output. Due to the lack of data to provide such supervision, we leverage the existing text-guided zero-shot segmentation model~\cite{luddecke2022image}, using its output as a pseudo label.}
  \vspace{-1em}
  \label{fig:fig2_2a}
\end{figure}

As shown in Fig.~\ref{fig:fig2_2a}, our audio-visual localizer produces a pixel-level probability mask given a sound input. This module is implemented based on a general encoder-decoder architecture. Our audio encoder is pre-trained jointly with the image-text CLIP embedding space (Sec.~\ref{sec:pretraining}), producing $d$-dimensional latent representation $\textbf{z}_a \in \mathbb{R}^d$. Given this $\textbf{z}_a$ and a latent representation for a source image $\textbf{z}_v$, our Transformer-based Audio-Visual Decoder is trained to produce a binary segmentation mask (Sec.~\ref{sec:audio-visual-decoder}).

\subsubsection{Pre-training Audio Encoder}\label{sec:pretraining} 
Inspired by Lee~\textit{et al.}~\cite{lee2022sound}, we first map the audio embeddings to the image-text CLIP embedding space via contrastive learning.
We adopt the InfoNCE loss~\cite{alayrac2020self} to map the same scene with different modalities close to each other while that of different scenes are far away in the CLIP embedding space~\cite{radford2021learning}.  
Furthermore, Lee~\textit{et al.} apply contrastive learning between audio latent representations as well as cross-modal. To generalize this technique, we interpolate the latent representations $\textbf{z}_{av} \in \mathbb{R}^d $ linearly between audio embedding $\textbf{z}_a  \in \mathbb{R}^d $ and visual embedding $\textbf{z}_v \in \mathbb{R}^d$ as 
\begin{equation}
    \begin{aligned}
        \textbf{z}_{av} = \alpha \cdot \textbf{z}_a + (1 - \alpha) \cdot \textbf{z}_v,
    \end{aligned}
    \label{eq:interpolation}
\end{equation}
where $\alpha$ is randomly sampled from $[0, 1]$. We employ the augmented audio input and image to obtain diverse latent representations $\hat{\textbf{z}}_a, \hat{\textbf{z}}_v \in \mathbb{R}^d$. In the same way, we obtain $\hat{\textbf{z}}_{av} \in \mathbb{R}^d $ with $\hat{\textbf{z}}_a$ and $\hat{\textbf{z}}_v$. 

Given mini-batches of size $N$ of interpolated latent representations, we minimize the InfoNCE loss with the interpolated latent representation pairs $\{\textbf{z}_{av}^i,\hat{\textbf{z}}_{av}^k\}$ for $i \in \{1,2...,N\}$ as follows:
\begin{equation}
    \begin{aligned}
        \mathcal{L}_\text{ctr}(i) = -\log{\exp{(\textbf{z}_{av}^i \cdot \hat{\textbf{z}}_{av}^i / \tau)}\over\sum_{k=1}^N \exp{(\textbf{z}_{av}^i \cdot \hat{\textbf{z}}_{av}^k / \tau)}},
    \end{aligned}
    \label{eq:clip_ccl}
\end{equation}
where $\tau$ is the temperature scale and $\cdot$ denotes the cosine similarity. Note that those latent representations are $l_2$-normalized. Then, we minimize the following loss term $\mathcal{L}_\text{ctr}$, defined in Eq.~\ref{eq:ccl_loss}, for all positive latent representation pairs in each mini-batch.
\begin{equation}
    \begin{aligned}
        \mathcal{L}_\text{ctr} = {1 \over N}\sum_{i=1}^N \mathcal{L}_\text{ctr}(i).
    \end{aligned}
    \label{eq:ccl_loss}
\end{equation}
For audio augmentation, we adopt the SpecAugment~\cite{park2019specaugment} that generates augmented Mel-spectrogram acoustic features. This augmentation strategy randomly warps the features and masks blocks in frequency channels.

\subsubsection{Audio-Visual Decoder}\label{sec:audio-visual-decoder}

\myparagraph{Weakly-Supervised Learning.} After pre-training the audio encoder, our audio-visual decoder learns to produce a pixel-level sound source location probability mask conditioned on the audio latent representation. Given audio, randomly sampled middle frame, and text descriptions $x_a$, $x_v$, $x_t$ from the video, we obtain $l_2$-normalized vector $\textbf{z}_a, \textbf{z}_t, \textbf{z}_v$ with the pre-trained audio encoder and CLIP image, text encoder. Then, we generate the pseudo ground-truth mask using text-based zero-shot segmentation, CLIPSeg~\cite{luddecke2022image}. Given the input image of width $W$ and height $H$, 
we compute the pseudo-segmentation mask $M_\text{pseudo} \in \mathbb{R}^{1 \times W \times H}$ as follows:
\begin{equation}
    M_\text{pseudo} =
      \begin{cases}
        1     & \sigma(\mathcal{F}_v(x_v, \textbf{z}_t)) > \text{threshold}, \\
        0  & \text{otherwise}, 
    \end{cases}
    \label{eq:mask_threshold}
\end{equation}
where $\sigma$ denotes the sigmoid function. CLIPSeg decoder $\mathcal{F}_v$ predicts the segmentation map $M_\text{pseudo}$ with text latent representation $\textbf{z}_t$ and the input image $x_v$. In the same way, audio-visual decoder $\mathcal{F}_a(\cdot, \cdot)$ outputs $M_\text{pred} \in \mathbb{R}^{1 \times W \times H}$ by $\mathcal{F}_a(x_v, \textbf{z}_a)$ conditioned on the input image $x_v$ and audio latent representation $\textbf{z}_a$ as follows:
\begin{equation}
    M_\text{pred} = \sigma(\mathcal{F}_a(x_a, \textbf{z}_a)).
    \label{eq:pred_mask}
\end{equation}

\myparagraph{Loss Function.} % Audio-visual localization decoder 
We minimize pixel-wise binary cross entropy loss between $M_\text{pseudo}$ and the predicted segmentation mask $M_\text{pred}$ as follows:
\begin{equation}
    \begin{aligned}
        \mathcal{L}_\text{bce} = {1 \over W \cdot H}\sum_i^W\sum_j^H M_\text{pseudo}^{(i,j)} \cdot \log{M_\text{pred}^{(i,j)}} \\ + (1 - M_\text{pseudo}^{(i,j)})\cdot\log{M_\text{pred}^{(i,j)}},
    \end{aligned}
    \label{eq:clip_ce}
\end{equation}
where $W, H$ denotes the width and height of the output probability mask and the pseudo mask.

\begin{figure*}[t!]
  \centering
  \includegraphics[width = \linewidth]{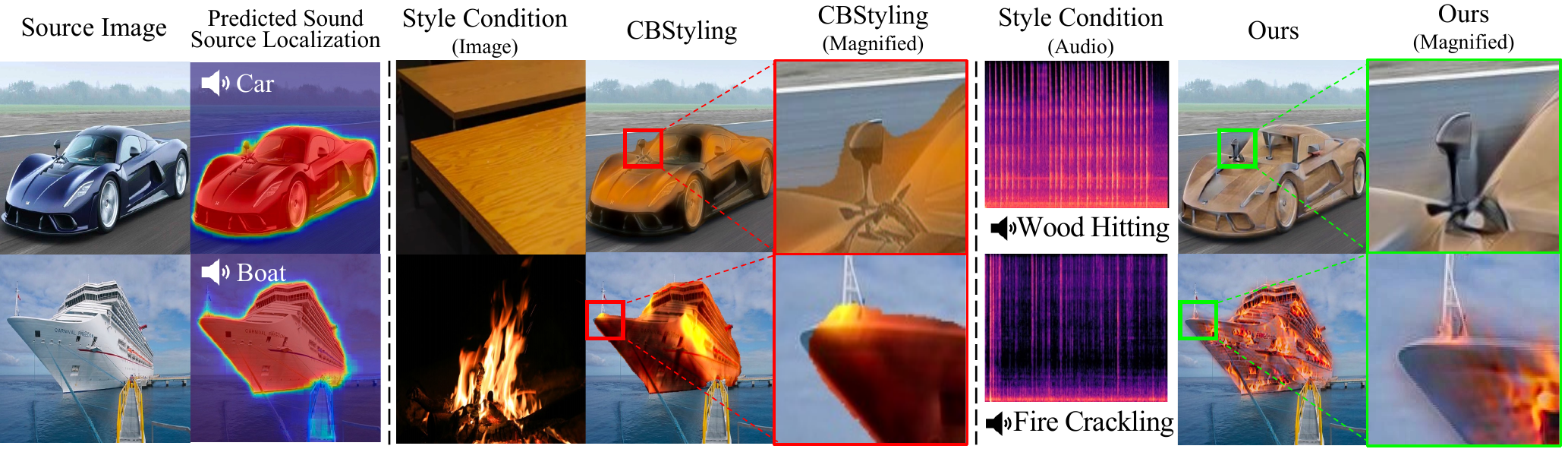}
  \caption{
   Comparison of our method to other existing localized neural style transfer, CBStyling~\cite{kurzman2019class} which both utilize localization mask. 
   For a fair comparison, we apply the same sound source localization map created with our method to CBStyling and ours.
   Our method produces a more plausible effect on the edge of the predicted mask than the baseline.
%   fully supervised approach.
  }
  \vspace{-1.0em}
  \label{fig:fig4}
\end{figure*}

\subsection{Audio-Guided INR Stylizer}
\label{sec:local}

To locally stylize an image based on the audio input, we introduce an audio-guided INR stylizer.
It consists of three modules: mask generator, INR, and multi-scale patching, shown in Fig.~\ref{fig:overview}. First, our mask generation process outputs the probability mask to determine the degree of style within the segmentation area of the sounding object using the pre-trained audio-visual localizer. 
Using the probability mask, the INR module predicts pixel-level continuous neural fields of style via multi-layer perceptron~(MLP). 
We also introduce the multi-scale patch sampling in CLIP embedding space to make the stylized image more realistic and dynamic.

\myparagraph{Mask Generation Process.} Given audio input $x_a$ and the source image $\mathcal{I}_x \in \mathbb{R}^{3\times W \times H}$, we first create a mask to stylize only a specific part of the source image. Audio latent representation $\textbf{z}_a$ is obtained by the pre-trained audio encoder. Then, we use the pre-trained audio-visual localizer~$\mathcal{F}_a$ to generate the probability mask $M \in \mathbb{R}^{1 \times W \times H}$ for sounding object or scene in the source image. $M$ is formulated as follows:
\begin{equation}
    M = \sigma(\mathcal{F}_a(\mathcal{I}_x, \textbf{z}_a)),
    \label{eq:pred_mask_final}
\end{equation}
where $\sigma$ denotes a sigmoid function.

\myparagraph{INR.} The goal of INR is to stylize the specific object or scene in the source image by using the predicted probability mask from the mask generator. 
Given the source image $\mathcal{I}_x$, the implicit neural network $f_\theta:\mathbb{R}^2 \rightarrow \mathbb{R}^3$ is optimized with parameters $\theta$ by mapping the pixel location $(i,j) \in \mathcal{I}_y$ into RGB pixel values $(r,g,b)$. Formally, the stylized image $\mathcal{I}_y$ is composited by the output of implicit neural representation $f_\theta$ and the pixel value of the source image as follows:
\begin{equation}
    \begin{aligned}
        \mathcal{I}_y(i,j) = M(i, j) \cdot f_\theta(\gamma(i,j)) + (1 - M(i, j)) \cdot \mathcal{I}_x(i, j).
    \end{aligned}
    \label{eq:image}
\end{equation}
Fourier feature mapping function $\gamma$ encodes the pixel coordinate to higher dimension space before feeding into MLP, which enhances high-frequency details~\cite{tancik2020fourier}.

\myparagraph{Multi-Scale Patch Sampling.} Given the source image and the INR output, We split the foreground~$ M(i, j) \cdot f_\theta(\gamma(i,j))$ and background~$(1 - M(i, j)) \cdot \mathcal{I}_x(i, j)$. The foreground is then used for multi-scale patching, and the background is copied to obtain the final stylized image. We crop the foreground into $K$ number of patches denoted by $\mathcal{P}_{k}$ for $k\in\{1,..,K\}$ where the pixel location of the patch center is in the segmentation mask with the predicted probability mask. 
The scale of each patch is selected randomly and each patch of the foreground is obtained by random augmentation strategy. 
We use these patches instead of the whole image to compute the similarity between audio and image embeddings in CLIP embedding space.
This multi-scale patch sampling helps to produce more dynamic and realistic style patterns as shown in the experiment section~(Fig.~\ref{fig:patch}).

\subsubsection{Training} 

For training, we use two losses: multi-scale PatchCLIP loss and foreground regularization loss. To stylize the source image with the audio input, we minimize the cosine distance in CLIP embedding space~\cite{radford2021learning} between the latent representation of audio and the multi-scale patches from the output of INR. Furthermore, we propose a foreground regularization loss to preserve the content of the source image by minimizing the pre-trained VGG feature distance between the source image and the stylized image. 

\myparagraph{Multi-Scale PatchCLIP Loss $\mathcal{L}_\text{CLIP}$.} Following Kwon~\textit{et al.}~\cite{kwon2022clipstyler}, we adopt the PatchCLIP loss to minimize cosine distance between randomly cropped patches and target audio in the CLIP embedding space. Given stylized patches~$\mathcal{P}_{k}$ and audio, we employ a CLIP-based visual and audio encoder to obtain visual and audio latent representation $\textbf{v}_k, \textbf{a} \in \mathbb{R}^d$. We minimize multi-scale PatchCLIP loss as follows:
\begin{equation}
    \begin{aligned}
        \mathcal{L}_\text{CLIP} = {1 \over K}\sum_{k=1}^K (1 - \langle\Delta \textbf{v}_k, \textbf{a}\rangle),
    \end{aligned}
    \label{eq:direc}
\end{equation}
where $K$ denotes the number of sampled patches. The visual direction $\Delta \textbf{v}$ is computed by subtracting the embedding of the $k$-th  cropped patch $\textbf{v}_k$ and the visual embedding of the source image. $\langle \Delta \textbf{v}_k, \textbf{a} \rangle$ represents the cosine similarity, i.e. $\langle \Delta \textbf{v}_k, \textbf{a} \rangle=\Delta \textbf{v}_k^\top\textbf{a}/||\Delta \textbf{v}_k||||\textbf{a}||$. Online hard example mining~(OHEM)~\cite{shrivastava2016training} is also applied to select a small number of hard examples, which avoids stylizing only the easy patches during optimization. Instead of minimizing the average of the cosine distances for each patch, we arrange the patches in the order of the largest cosine distance and then minimize the number of patches with large distances.

\myparagraph{Foreground Regularization Loss $\mathcal{L}_\text{reg}$.} 
Preserving the source image is crucial for local stylization. 
To obtain a perceptually reasonable stylized image, we apply perceptual loss~\cite{johnson2016perceptual} to the entire image as follows:
\begin{equation}
	\begin{split}
	\mathcal{L}_\text{reg} = \sum_{k=0}^K |G_k^\phi(\mathcal{I}_y)- G_k^\phi(\mathcal{I}_x)|,
	\end{split}
	\label{eq:vgg}
\end{equation}
where $G_k^\phi$ denotes the gram matrix of $k$-th layer's respective feature maps from the pre-trained VGG-16 network.
We minimize the $L_1$ loss between the gram matrix of the source image and that of the stylized image.

\myparagraph{Total Loss.} Ultimately, we minimize the following loss function $\mathcal{L}_\text{total}$:
\begin{equation}
    \begin{aligned}
        \mathcal{L}_\text{total} = \lambda_\text{CLIP}\mathcal{L}_\text{CLIP} + \lambda_\text{reg}\mathcal{L}_\text{reg} + \lambda_\text{c}\mathcal{L}_\text{c},
    \end{aligned}
    \label{eq:final}
\end{equation}
where $\lambda_\text{CLIP}, \lambda_\text{reg},$ and $\lambda_\text{c}$ are hyperparameters to tune the strength of each term. We also apply content loss function $\mathcal{L}_\text{c}$ following Johnson~\textit{et al.}~\cite{7780634}, which helps preserve the visual properties of the source image by minimizing the mean squared error between the features of the source and the output images with the pre-trained VGG-19~\cite{gatys2015neural}.

\begin{figure}[t!]
  \centering
  \includegraphics[width = \linewidth]{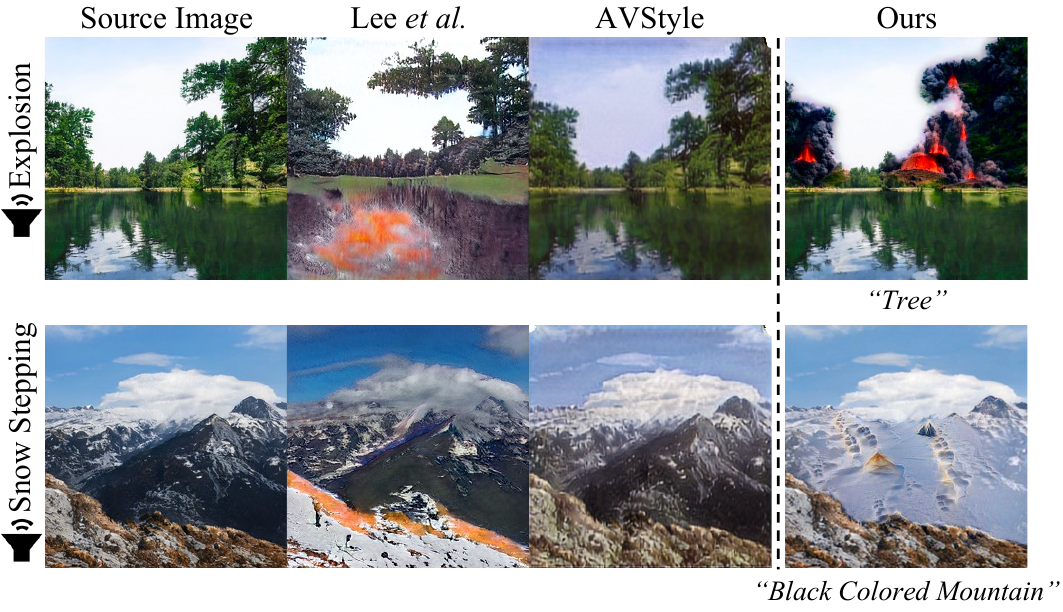}
  \vspace{-2em}
  \caption{
    Qualitative comparison with existing audio-guided image stylization methods, including Lee~\textit{et al.}~\cite{lee2022sound} and AVStyle~\cite{li2021learning}. Unlike existing approaches, our model localizes visual attributes corresponding to a text or sound input, enabling a localized stylization (see ours stylizes regions corresponding to ``Tree'' or ``Black Colored Mountain'').
  }
  \label{fig:fig5}
  % \vspace{-1.5em}
\end{figure}

\section{Experiments}

\subsection{Implementation Details}

\myparagraph{Datasets.} We composite two different dataset for pre-training audio encoder and decoder:~\textit{Greatest Hits}~\cite{owens2016visually} and~\textit{VGG-Sound}~\cite{chen2020vggsound} for mapping CLIP embedding space~\cite{radford2021learning} with audio latent representation. \textit{VGG-Sound} contains large-scale audio-visual pairs.
The number of audio-visual pairs in the~\textit{VGG-Sound} dataset is over 200K. However, we collect 155,040 videos for training due to missing video clips on YouTube.
The \textit{Greatest Hits} dataset includes various audios of hitting and scratching various objects with a drumstick. The number of the \textit{Greatest Hits} training dataset is 733 videos in total. Note that there is no ground-truth for the segmentation mask in the composited dataset. 

\myparagraph{Training Details of Audio-Visual Localizer.} To represent the audio latent vector, we adopt Swin-S~\cite{liu2021Swin} architecture as the audio encoder backbone and the output dimension of the audio encoder is 512. 
Audio-visual decoder follows the CLIPSeg~\cite{luddecke2022image} architecture. We employ the (ViT-B/16)~CLIP transformer-based decoder~\cite{radford2021learning}. The input image size is fixed as 352$\times$352 for training the audio-visual localizer. 
We train the audio-visual decoder for 20 epochs with Adam optimizer. We choose the cosine cyclic learning rate scheduler. The batch size for training is determined as 160.

\begin{figure}[t!]
  \centering
  \includegraphics[width = \linewidth]{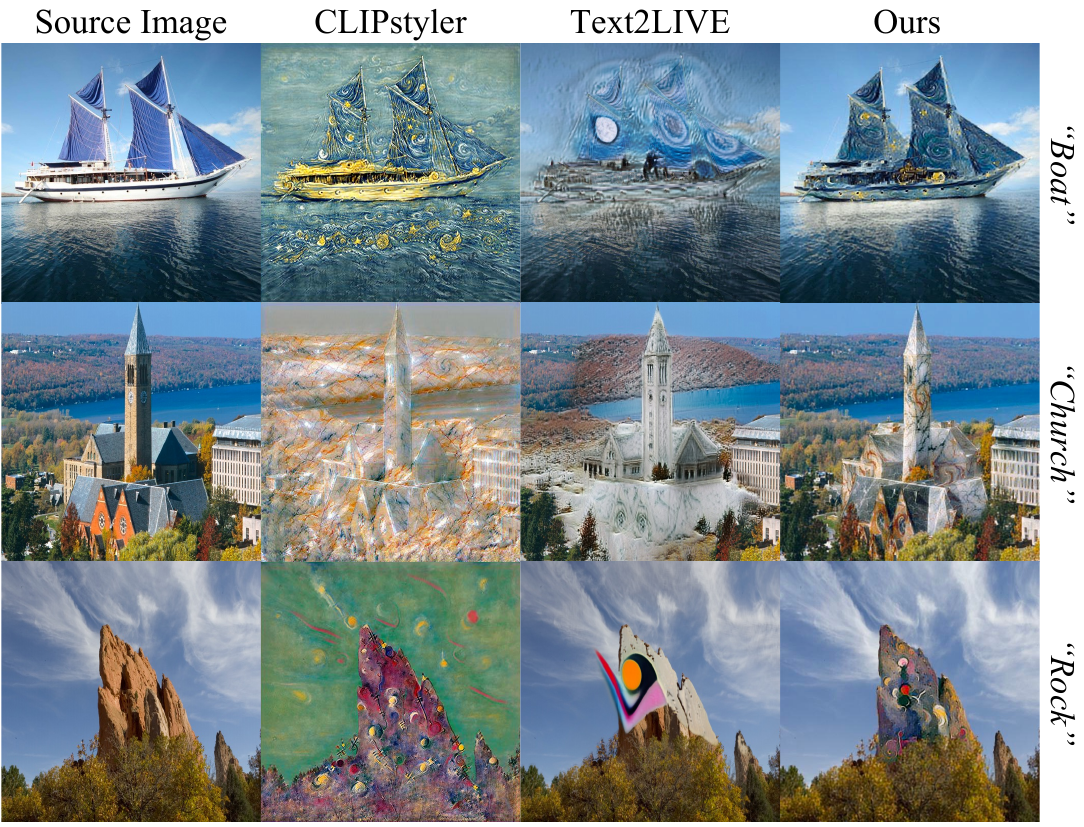}
  \caption{
    Stylization output comparison with existing text-driven stylization approaches (CLIPstyler~\cite{kwon2022clipstyler} and Text2LIVE~\cite{bar2022text2live}). We set \textit{The Starry Night by Vincent Van Gogh}, \textit{Marble}, and \textit{Composition VII by Wassily Kandinsky} as style conditions (respectively, from top to bottom).
    }
  \label{fig:fig3}
  \vspace{-0.5em}
\end{figure}

\myparagraph{INR Architecture.} We follow the basic structure of implicit neural representation instead of the conventional U-Net-based approach. 
Based on Tancik~\textit{et al.}~\cite{tancik2020fourier}, we replace ReLU activation with SIREN~\cite{sitzmann2020implicit}.
The parameters of MLP~(8 layers, 256 channels, SIREN activation) are trained with Fourier feature mappings. 

\myparagraph{Hyperparameter Setup.} We use 512$\times$512 resolution for training. For localized style transfer, we remove the global CLIP loss which is defined in CLIPStyler~\cite{kwon2022clipstyler}. The loss weight $\lambda_\text{CLIP},~\lambda_\text{reg},~\lambda_\text{c}$ is set to 35, 0.2, 2. We randomly select the multi-scale patch sizes in the range $[64,256]$.

\subsection{Comparision to Baselines}
To the best of our knowledge, our work is the first to attempt audio-driven localized stylization of sounding objects from the wild image. We compare our method to several state-of-the-art image stylization methods carefully. 

\myparagraph{Localized Style Transfer.} To compare our method to a mask-based style transfer model, we choose CBStyling~\cite{kurzman2019class} as a baseline. 
Fig.~\ref{fig:fig4} shows the qualitative comparison of our method to the previous approach. To obtain mask, our method uses the audio input as a condition and generates the mask using audio-visual localizer. However, CBStyling requires reference style image and mask with full class supervision. For fairness, we report the CBStyling result with the same sound source localization map. We observe that our proposed model produces more realistic stylized image than CBStyling on the edge of the predicted mask. This is because the previous blending approaches simply crop the result of the stylization using mask and paste it into the source image. However, our INR-based method jointly optimizes the stylization loss and foreground regularization loss.

\myparagraph{Audio-Guided Image Stylization.} We compare our model to the existing audio-driven image stylization methods. We select two baselines: Lee~\textit{et al.}~\cite{lee2022sound} and AVStyle~\cite{li2021learning}. As Fig.~\ref{fig:fig5} shows, none of the existing audio-guided image stylization methods perform localized image stylization. Although Lee~\textit{et al.} changes specific parts of scenes with semantic cues from audio, the area of stylization is not controllable~(see $2^\text{nd}$ row, $3^\text{rd}$ column). Furthermore, Lee~\textit{et al.} fails to maintain some geometric features, such as the shape of tree branches~(see $1^\text{st}$ row, $3^\text{rd}$ column). Note that Lee~\textit{et al.} cannot manipulate diverse wild image domains because their work needs the mapping function of the input image to the latent space of the pre-trained generative model. AVStyle fails to stylize the image with out-of-domain audio~(e.g., explosion), which is not included in the audio-visual training dataset. In contrast, our method can cover various audio by leveraging CLIP embedding space~\cite{radford2021learning}.

\myparagraph{Interactive Stylization with Audio \& Text.} We compare our style transfer model to state-of-the-art text-driven image editing models, CLIPStyler~\cite{kwon2022clipstyler} and Text2LIVE~\cite{bar2022text2live}. Note that our model can perform image stylization under various audio and text conditions because our audio latent representation shares with CLIP embedding space. Fig.~\ref{fig:fig3} demonstrates that our model produces more reasonable stylization results at the object boundary than Text2LIVE. This is because our method explicitly provides the predicted probability mask as a condition. We further show more examples in the supplementary material.

\begin{figure}[t!]
  \centering
  \includegraphics[width = \linewidth]{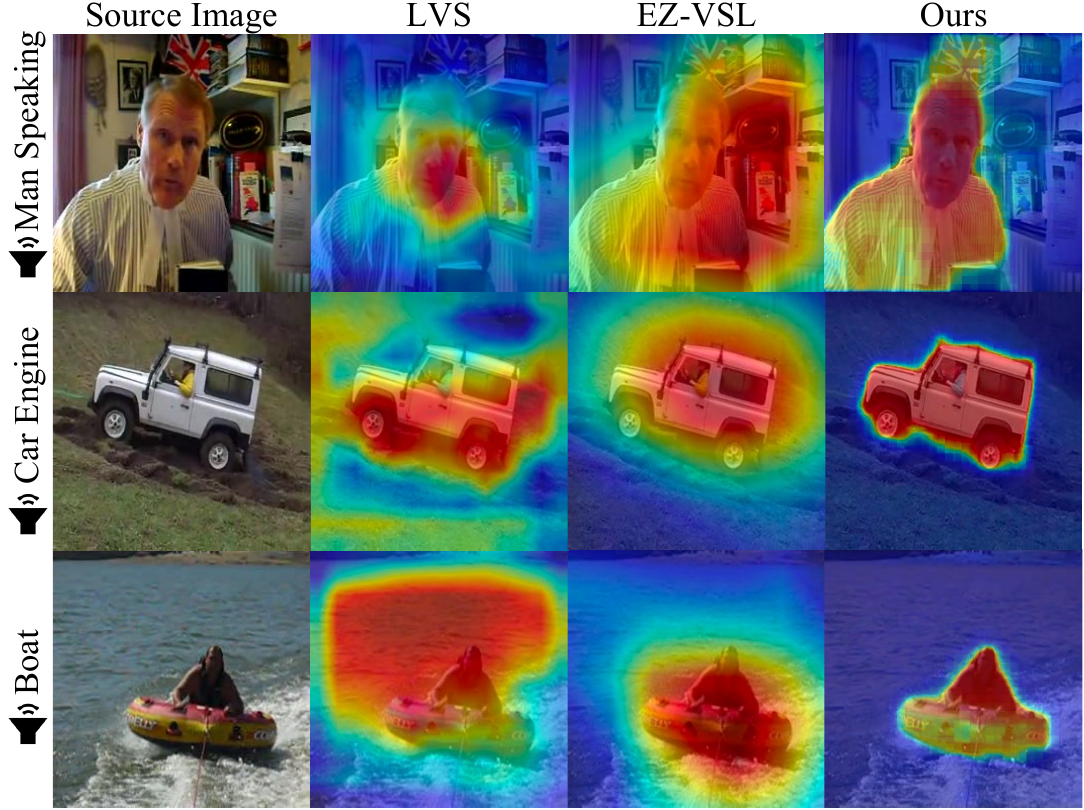}
  \caption{
   Localization quality comparison with existing state-of-the-art sound-based visual localization methods, including LVS~\cite{chen2021localizing} and EZ-VSL~\cite{10.1007/978-3-031-19836-6_13}. For each method, we visualize a predicted heatmap of sound-based localization.
  }
  \label{fig:fig_loc}
  \vspace{-0.5em}
\end{figure}

\subsection{Audio-Visual Localizer}
\myparagraph{Qualitative Analysis.} In qualitative comparisons, we visualize the localization with a state-of-the-art audio-visual source localization model, EZ-VSL~\cite{10.1007/978-3-031-19836-6_13} which is pre-trained with the full VGG-Sound~\cite{chen2020vggsound} dataset. 
We sample images from the Flickr SoundNet sound localization dataset~\cite{senocak2018learning} to qualitatively evaluate audio-visual source localization. 
% As Fig.~\ref{fig:fig_loc} shows, our model localizes objects better than the other baselines through a heatmap.
We qualitatively demonstrate our audio-visual localization method by visualizing the heatmap in Fig.~\ref{fig:fig_loc}.
We observe that our model has the capability of producing a more precise audio-visual source localization performance compared to other baselines.

\myparagraph{Quantitative Evaluation.} In Table.~\ref{table:segmentation}, we quantitatively compare our audio-visual source localization with existing audio-visual source localization methods. We choose LVS~\cite{chen2021localizing} and EZ-VSL~\cite{10.1007/978-3-031-19836-6_13} as baselines for audio-visual source localization. Baseline models have trained in an unsupervised manner with the full VGG-Sound~\cite{chen2020vggsound} dataset. We choose the Flickr SoundNet testset~\cite{senocak2018learning} for quantitative evaluation. We measure the two metrics for sound source localization, Consensus Interaction over Union~(CIoU), and the Area Under Curve~(AUC). We set a CIoU threshold as 0.5 for all baselines.

\begin{table}[t!]
    \caption{Quantitative comparison between the existing audio-visual sound source localization methods and ours. We report CIoU and AUC for sound source localization~(SSL).
    }
    \label{table:segmentation}
    \centering
    % \resizebox{\linewidth}{!}{
    \vspace{-0.5em}
    \small
    \begin{tabular}{@{}lcc@{}}
    
        \toprule
        \multirow{2}{*}{Model} & \multicolumn{2}{c}{SSL Metric}  \\ \cmidrule{2-3}
        & CIoU~($\uparrow$) & AUC~($\uparrow$) \\ 
        \midrule
        LVS~\cite{chen2021localizing} & 73.59 \% & 59.00 \%  \\ 
        EZ-VSL \cite{10.1007/978-3-031-19836-6_13} &  83.94 \% & 63.60 \% \\ 
        \midrule
        Ours & \textbf{85.94} \% & \textbf{66.70} \%  \\
        \bottomrule
    \end{tabular}
\end{table}

\begin{figure}[t!]
  \centering
  \includegraphics[width = \linewidth]{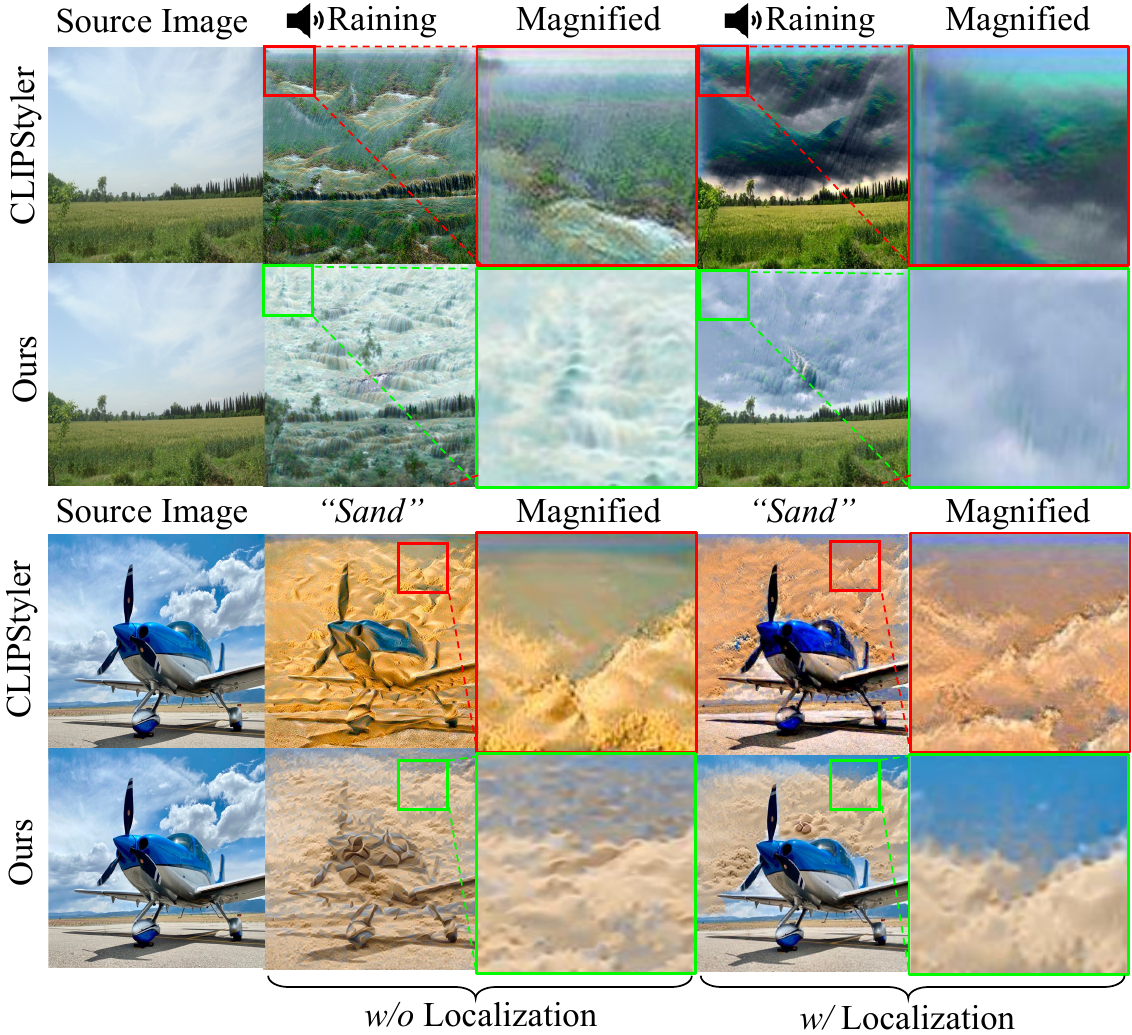}
  \caption{
   Stylization quality comparison between conventional U-Net-based~\cite{kwon2022clipstyler} vs. INR-based approach (ours). Note that the former suffers from artifacts near borders, while the INR-based approach generates clearer images. 
  }
  \vspace{-0.5em}
  \label{fig:fig7_1}
\end{figure}

\subsection{Ablation Study}

\myparagraph{Effect of INR in Style Transfer.} Fig.~\ref{fig:fig7_1} demonstrates that implicit neural representations are effective in artifact removal for local stylization. 
The first row is the result of stylization with audio~\textit{Thunderstorm}, and the second row is the~\textit{airplane} image stylized with text condition~\textit{Desert}. 
The downscaling and upscaling operation with zero-padding in the U-Net-based style transfer~\cite{kwon2022clipstyler} loses fine-level visual details and generates line-shaped border artifacts~\cite{nguyen2019distribution} at the edge of the stylized image.
On the contrary, our stylized result produces a clean image without any disturbance of the perceptual naturalness since our INR maps $(x, y)$ coordinate directly into an RGB pixel value without affecting neighboring pixels.

\begin{figure}[t!]
  \centering
  \includegraphics[width = \linewidth]{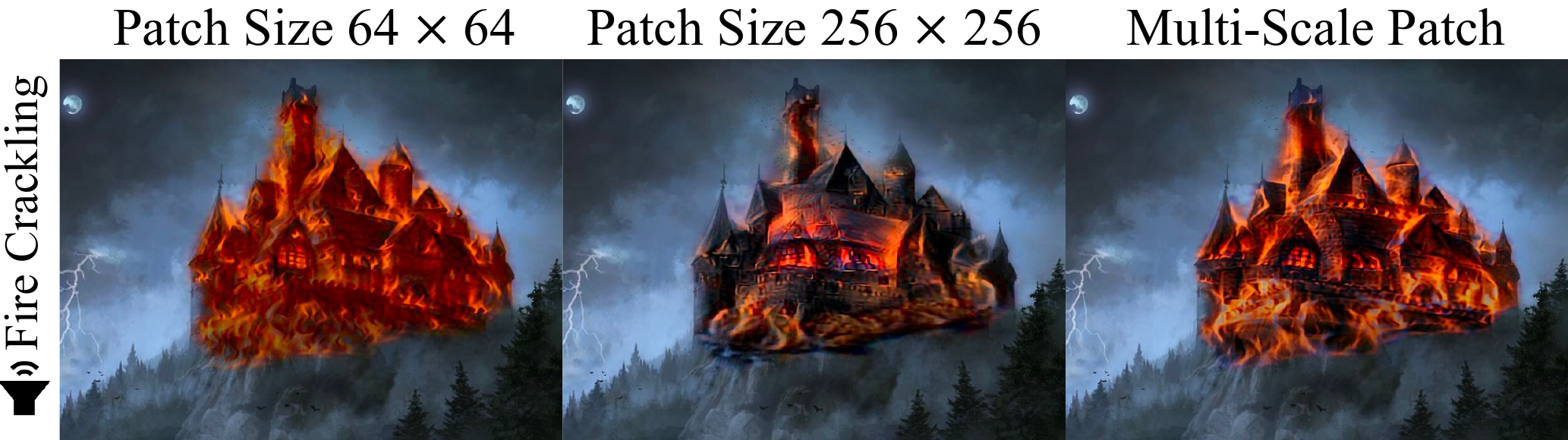}
  \caption{
   Effect of our Multi-scale PatchCLIP loss compared to a model with Single-scale PatchCLIP loss given a fixed patch size.
  }
  \label{fig:patch}
  \vspace{-0.5em}
\end{figure}

% Therefore we use 8 MLP

\myparagraph{Multi-Scale PatchCLIP Loss.} Fig.~\ref{fig:patch} shows the stylization results according to multi-scale PatchCLIP loss. We compare by increasing the patch size as follows: 64$\times$64, 256$\times$256, and multi-scale patches. We observe that a larger patch size results in a larger stylization area in the source image. By using multi-scale patches, we can produce more dynamic and realistic style patterns.
% Fig.~\ref{fig:fig8} 

\myparagraph{The Number of SIREN Layers.} We perform the ablation study about the number of fully-connected SIREN~\cite{sitzmann2020implicit} layers. We compare the stylization results by increasing the number of SIREN layers to 2, 8, 14, and 20. As the number of layers increases, the style change becomes finer, and more iterations are required for convergence, which results in more time consumption~(see the supplemental material).

\subsection{User Study}
We gather 100 participants from Amazon Mechanical Turk to assess whether our proposed method is realistic from a human perspective. 
Participants are asked 20 questions with binary choices to validate our proposed method.
We investigate two factors, including~\textit{Naturalness} and~\textit{Attribute Consistency}. Participants are required to answer binary choice questions to assess our method against the other method directly.

\begin{figure}[t!]
  \centering
  \includegraphics[width = \linewidth]{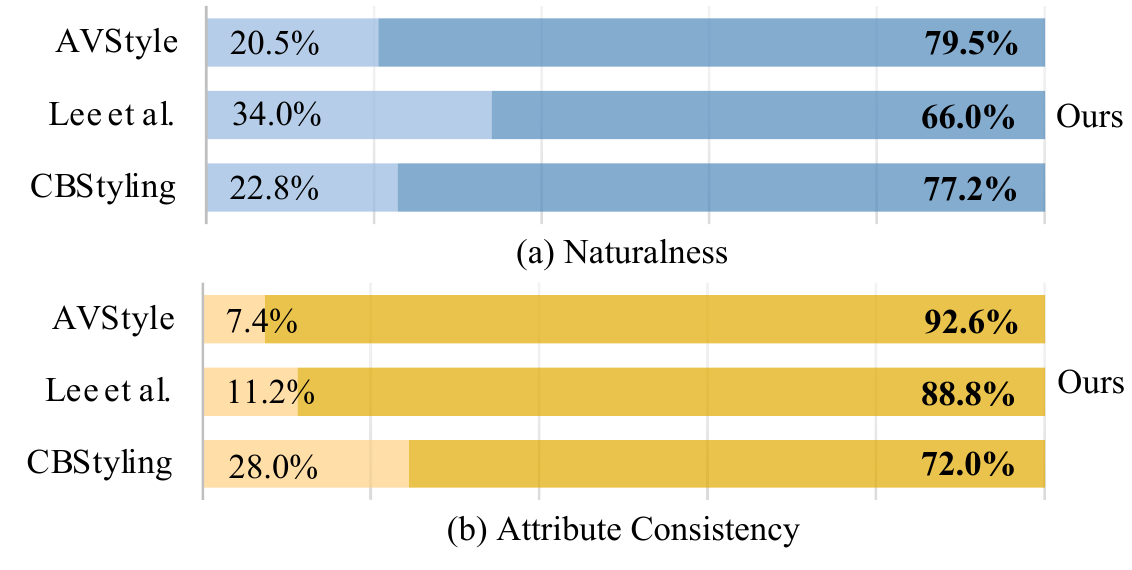}
  \vspace{-2em}
  \caption{
    User study results comparing LISA vs. other baselines. Human evaluators are asked to complete the questionnaire about (a) Naturalness~\textit{``Which stylization result is more natural, especially at the edge of each target object?''} and (b) Attribute Consistency ~\textit{``Which image is better stylized in the object localization?''}.
  }
  \label{fig:user}
  \vspace{-0.5em}
\end{figure}

Since AVStyle~\cite{li2021learning} and Lee~\textit{et al.}~\cite{lee2022sound} do not have localization capability, we stylize the global region of the source image.
For a fair comparison, we give CBStyling~\cite{kurzman2019class} the same sound source localization map as our model because CBStyling uses a different segmentation network from ours.
Fig.~\ref{fig:user} shows that our method excels the other state-of-the-art methods in terms of stylization naturalness~($69.83\%$ in average) and attribute consistency~($84.47\%$ in average).

\section{Discussion}

\myparagraph{Limitation.} 
The implicit function offers fast training time compared to existing CNN-based models. However, since the CLIP loss is a bottleneck, the overall optimization speed is similar to that of the CLIPStyler.

\myparagraph{Broader Impact.} We can stylize user-desired parts of the source image with the audio input. Since our model stylizes images solely on user intention, the images that the user stylized can raise ethical concerns or may apply a style that is opposite to the intention of the original image creator.

\section{Conclusion}
In this work, we propose a novel framework for audio-guided local image stylization, named \textit{LISA}. Audio-visual sound source localizer provides a delicate localization map by leveraging the CLIP embedding space in a weakly supervised manner.
Using this localization map, the coordinate-based implicit neural representation effectively produces more realistic and vivid stylization results than cut-and-paste approaches. We also observe that our network helps to remove line-shaped border artifacts that existed in CNN-style mapping structures. 
Moreover, our method is able to perform text-driven local and semantic stylization as well as audio because our audio latent representation shares CLIP embedding space.

\newpage

%%%%%%%%% REFERENCES
{\small
\bibliographystyle{ieee_fullname}
\bibliography{egbib}

\begin{thebibliography}{10}\itemsep=-1pt

\bibitem{alayrac2020self}
Jean-Baptiste Alayrac, Adria Recasens, Rosalia Schneider, Relja
  Arandjelovi{\'c}, Jason Ramapuram, Jeffrey De~Fauw, Lucas Smaira, Sander
  Dieleman, and Andrew Zisserman.
\newblock Self-supervised multimodal versatile networks.
\newblock {\em Advances in Neural Information Processing Systems}, 33:25--37,
  2020.

\bibitem{alegre2020selfieart}
Lucas~N Alegre and Manuel~M Oliveira.
\newblock Selfieart: Interactive multi-style transfer for selfies and videos
  with soft transitions.
\newblock In {\em 2020 33rd SIBGRAPI Conference on Graphics, Patterns and
  Images (SIBGRAPI)}, pages 17--22. IEEE, 2020.

\bibitem{bar2022text2live}
Omer Bar-Tal, Dolev Ofri-Amar, Rafail Fridman, Yoni Kasten, and Tali Dekel.
\newblock Text2live: Text-driven layered image and video editing.
\newblock In Shai Avidan, Gabriel Brostow, Moustapha Ciss{\'e}, Giovanni~Maria
  Farinella, and Tal Hassner, editors, {\em Computer Vision -- ECCV 2022},
  pages 707--723, Cham, 2022. Springer Nature Switzerland.

\bibitem{castillo2017zorn}
Carlos Castillo, Soham De, Xintong Han, Bharat Singh, Abhay~Kumar Yadav, and
  Tom Goldstein.
\newblock Son of zorn's lemma: Targeted style transfer using instance-aware
  semantic segmentation.
\newblock In {\em 2017 IEEE International Conference on Acoustics, Speech and
  Signal Processing (ICASSP)}, pages 1348--1352. IEEE, 2017.

\bibitem{chen2021localizing}
Honglie Chen, Weidi Xie, Triantafyllos Afouras, Arsha Nagrani, Andrea Vedaldi,
  and Andrew Zisserman.
\newblock Localizing visual sounds the hard way.
\newblock In {\em Proceedings of the IEEE/CVF Conference on Computer Vision and
  Pattern Recognition}, pages 16867--16876, 2021.

\bibitem{chen2020vggsound}
Honglie Chen, Weidi Xie, Andrea Vedaldi, and Andrew Zisserman.
\newblock Vggsound: A large-scale audio-visual dataset.
\newblock In {\em ICASSP 2020-2020 IEEE International Conference on Acoustics,
  Speech and Signal Processing (ICASSP)}, pages 721--725. IEEE, 2020.

\bibitem{chen2018lip}
Lele Chen, Zhiheng Li, Ross~K Maddox, Zhiyao Duan, and Chenliang Xu.
\newblock Lip movements generation at a glance.
\newblock In {\em Proceedings of the European Conference on Computer Vision
  (ECCV)}, pages 520--535, 2018.

\bibitem{chiang2022stylizing}
Pei-Ze Chiang, Meng-Shiun Tsai, Hung-Yu Tseng, Wei-Sheng Lai, and Wei-Chen
  Chiu.
\newblock Stylizing 3d scene via implicit representation and hypernetwork.
\newblock In {\em Proceedings of the IEEE/CVF Winter Conference on Applications
  of Computer Vision}, pages 1475--1484, 2022.

\bibitem{deng2021unsupervised}
Kangle Deng, Aayush Bansal, and Deva Ramanan.
\newblock Unsupervised audiovisual synthesis via exemplar autoencoders.
\newblock In {\em International Conference on Learning Representations}, 2021.

\bibitem{gatys2015neural}
Leon~A Gatys, Alexander~S Ecker, and Matthias Bethge.
\newblock A neural algorithm of artistic style.
\newblock {\em arXiv preprint arXiv:1508.06576}, 2015.

\bibitem{7780634}
Leon~A. Gatys, Alexander~S. Ecker, and Matthias Bethge.
\newblock Image style transfer using convolutional neural networks.
\newblock In {\em 2016 IEEE Conference on Computer Vision and Pattern
  Recognition (CVPR)}, pages 2414--2423, 2016.

\bibitem{gatys2017controlling}
Leon~A Gatys, Alexander~S Ecker, Matthias Bethge, Aaron Hertzmann, and Eli
  Shechtman.
\newblock Controlling perceptual factors in neural style transfer.
\newblock In {\em Proceedings of the IEEE conference on computer vision and
  pattern recognition}, pages 3985--3993, 2017.

\bibitem{genova2020local}
Kyle Genova, Forrester Cole, Avneesh Sud, Aaron Sarna, and Thomas Funkhouser.
\newblock Local deep implicit functions for 3d shape.
\newblock In {\em Proceedings of the IEEE/CVF Conference on Computer Vision and
  Pattern Recognition}, pages 4857--4866, 2020.

\bibitem{genova2019learning}
Kyle Genova, Forrester Cole, Daniel Vlasic, Aaron Sarna, William~T Freeman, and
  Thomas Funkhouser.
\newblock Learning shape templates with structured implicit functions.
\newblock In {\em Proceedings of the IEEE/CVF International Conference on
  Computer Vision}, pages 7154--7164, 2019.

\bibitem{hollein2022stylemesh}
Lukas H{\"o}llein, Justin Johnson, and Matthias Nie{\ss}ner.
\newblock Stylemesh: Style transfer for indoor 3d scene reconstructions.
\newblock In {\em Proceedings of the IEEE/CVF Conference on Computer Vision and
  Pattern Recognition}, pages 6198--6208, 2022.

\bibitem{huang2017arbitrary}
Xun Huang and Serge Belongie.
\newblock Arbitrary style transfer in real-time with adaptive instance
  normalization.
\newblock In {\em Proceedings of the IEEE international conference on computer
  vision}, pages 1501--1510, 2017.

\bibitem{huang2022stylizednerf}
Yi-Hua Huang, Yue He, Yu-Jie Yuan, Yu-Kun Lai, and Lin Gao.
\newblock Stylizednerf: consistent 3d scene stylization as stylized nerf via
  2d-3d mutual learning.
\newblock In {\em Proceedings of the IEEE/CVF Conference on Computer Vision and
  Pattern Recognition}, pages 18342--18352, 2022.

\bibitem{johnson2016perceptual}
Justin Johnson, Alexandre Alahi, and Li Fei-Fei.
\newblock Perceptual losses for real-time style transfer and super-resolution.
\newblock In {\em European conference on computer vision}, pages 694--711.
  Springer, 2016.

\bibitem{kurzman2019class}
Lironne Kurzman, David Vazquez, and Issam Laradji.
\newblock Class-based styling: Real-time localized style transfer with semantic
  segmentation.
\newblock In {\em Proceedings of the IEEE/CVF International Conference on
  Computer Vision Workshops}, pages 0--0, 2019.

\bibitem{kwon2022clipstyler}
Gihyun Kwon and Jong~Chul Ye.
\newblock Clipstyler: Image style transfer with a single text condition.
\newblock In {\em Proceedings of the IEEE/CVF Conference on Computer Vision and
  Pattern Recognition}, pages 18062--18071, 2022.

\bibitem{lee2022robust}
Seung~Hyun Lee, Chanyoung Kim, Wonmin Byeon, Gyeongrok Oh, Jooyoung Lee,
  Sang~Ho Yoon, Jinkyu Kim, and Sangpil Kim.
\newblock Robust sound-guided image manipulation.
\newblock {\em arXiv preprint arXiv:2208.14114}, 2022.

\bibitem{lee2022sound}
Seung~Hyun Lee, Wonseok Roh, Wonmin Byeon, Sang~Ho Yoon, Chanyoung Kim, Jinkyu
  Kim, and Sangpil Kim.
\newblock Sound-guided semantic image manipulation.
\newblock In {\em Proceedings of the IEEE/CVF Conference on Computer Vision and
  Pattern Recognition}, pages 3377--3386, 2022.

\bibitem{li2021learning}
Tingle Li, Yichen Liu, Andrew Owens, and Hang Zhao.
\newblock {Learning Visual Styles from Audio-Visual Associations}.
\newblock In {\em European Conference on Computer Vision (ECCV)}, 2022.

\bibitem{liu2021Swin}
Ze Liu, Yutong Lin, Yue Cao, Han Hu, Yixuan Wei, Zheng Zhang, Stephen Lin, and
  Baining Guo.
\newblock Swin transformer: Hierarchical vision transformer using shifted
  windows.
\newblock In {\em Proceedings of the IEEE/CVF International Conference on
  Computer Vision (ICCV)}, 2021.

\bibitem{luan2017deep}
Fujun Luan, Sylvain Paris, Eli Shechtman, and Kavita Bala.
\newblock Deep photo style transfer.
\newblock In {\em Proceedings of the IEEE conference on computer vision and
  pattern recognition}, pages 4990--4998, 2017.

\bibitem{luddecke2022image}
Timo L{\"u}ddecke and Alexander Ecker.
\newblock Image segmentation using text and image prompts.
\newblock In {\em Proceedings of the IEEE/CVF Conference on Computer Vision and
  Pattern Recognition}, pages 7086--7096, 2022.

\bibitem{10.1007/978-3-031-19836-6_13}
Shentong Mo and Pedro Morgado.
\newblock Localizing visual sounds the easy way.
\newblock In Shai Avidan, Gabriel Brostow, Moustapha Ciss{\'e}, Giovanni~Maria
  Farinella, and Tal Hassner, editors, {\em Computer Vision -- ECCV 2022},
  pages 218--234, Cham, 2022. Springer Nature Switzerland.

\bibitem{mu20223d}
Fangzhou Mu, Jian Wang, Yicheng Wu, and Yin Li.
\newblock 3d photo stylization: Learning to generate stylized novel views from
  a single image.
\newblock In {\em Proceedings of the IEEE/CVF Conference on Computer Vision and
  Pattern Recognition}, pages 16273--16282, 2022.

\bibitem{nguyen2019distribution}
Anh-Duc Nguyen, Seonghwa Choi, Woojae Kim, Sewoong Ahn, Jinwoo Kim, and
  Sanghoon Lee.
\newblock Distribution padding in convolutional neural networks.
\newblock In {\em 2019 IEEE International Conference on Image Processing
  (ICIP)}, pages 4275--4279. IEEE, 2019.

\bibitem{owens2016visually}
Andrew Owens, Phillip Isola, Josh McDermott, Antonio Torralba, Edward~H
  Adelson, and William~T Freeman.
\newblock Visually indicated sounds.
\newblock In {\em Proceedings of the IEEE conference on computer vision and
  pattern recognition}, pages 2405--2413, 2016.

\bibitem{park2019specaugment}
Daniel~S Park, William Chan, Yu Zhang, Chung-Cheng Chiu, Barret Zoph, Ekin~D
  Cubuk, and Quoc~V Le.
\newblock Specaugment: A simple data augmentation method for automatic speech
  recognition.
\newblock {\em arXiv preprint arXiv:1904.08779}, 2019.

\bibitem{park2019deepsdf}
Jeong~Joon Park, Peter Florence, Julian Straub, Richard Newcombe, and Steven
  Lovegrove.
\newblock Deepsdf: Learning continuous signed distance functions for shape
  representation.
\newblock In {\em Proceedings of the IEEE/CVF conference on computer vision and
  pattern recognition}, pages 165--174, 2019.

\bibitem{radford2021learning}
Alec Radford, Jong~Wook Kim, Chris Hallacy, Aditya Ramesh, Gabriel Goh,
  Sandhini Agarwal, Girish Sastry, Amanda Askell, Pamela Mishkin, Jack Clark,
  et~al.
\newblock Learning transferable visual models from natural language
  supervision.
\newblock In {\em International Conference on Machine Learning}, pages
  8748--8763. PMLR, 2021.

\bibitem{senocak2018learning}
Arda Senocak, Tae-Hyun Oh, Junsik Kim, Ming-Hsuan Yang, and In~So Kweon.
\newblock Learning to localize sound source in visual scenes.
\newblock In {\em Proceedings of the IEEE Conference on Computer Vision and
  Pattern Recognition}, pages 4358--4366, 2018.

\bibitem{shrivastava2016training}
Abhinav Shrivastava, Abhinav Gupta, and Ross Girshick.
\newblock Training region-based object detectors with online hard example
  mining.
\newblock In {\em Proceedings of the IEEE conference on computer vision and
  pattern recognition}, pages 761--769, 2016.

\bibitem{sitzmann2020implicit}
Vincent Sitzmann, Julien Martel, Alexander Bergman, David Lindell, and Gordon
  Wetzstein.
\newblock Implicit neural representations with periodic activation functions.
\newblock {\em Advances in Neural Information Processing Systems},
  33:7462--7473, 2020.

\bibitem{smith2017cyclical}
Leslie~N Smith.
\newblock Cyclical learning rates for training neural networks.
\newblock In {\em 2017 IEEE winter conference on applications of computer
  vision (WACV)}, pages 464--472. IEEE, 2017.

\bibitem{son2017lip}
Joon Son~Chung, Andrew Senior, Oriol Vinyals, and Andrew Zisserman.
\newblock Lip reading sentences in the wild.
\newblock In {\em Proceedings of the IEEE conference on computer vision and
  pattern recognition}, pages 6447--6456, 2017.

\bibitem{stahl2019style}
Fabian Stahl, Melina Meyer, and Ulrich Schwanecke.
\newblock Ist-style transfer with instance segmentation.
\newblock In {\em 2019 11th International Symposium on Image and Signal
  Processing and Analysis (ISPA)}, pages 277--281. IEEE, 2019.

\bibitem{tancik2020fourier}
Matthew Tancik, Pratul Srinivasan, Ben Mildenhall, Sara Fridovich-Keil, Nithin
  Raghavan, Utkarsh Singhal, Ravi Ramamoorthi, Jonathan Barron, and Ren Ng.
\newblock Fourier features let networks learn high frequency functions in low
  dimensional domains.
\newblock {\em Advances in Neural Information Processing Systems},
  33:7537--7547, 2020.

\bibitem{tian2018audio}
Yapeng Tian, Jing Shi, Bochen Li, Zhiyao Duan, and Chenliang Xu.
\newblock Audio-visual event localization in unconstrained videos.
\newblock In {\em Proceedings of the European Conference on Computer Vision
  (ECCV)}, pages 247--263, 2018.

\bibitem{virtusio2018interactive}
John~Jethro Virtusio, Arces Talavera, Daniel~Stanley Tan, Kai-Lung Hua, and
  Arnulfo Azcarraga.
\newblock Interactive style transfer: Towards styling user-specified object.
\newblock In {\em 2018 IEEE Visual Communications and Image Processing (VCIP)},
  pages 1--4. IEEE, 2018.

\bibitem{xia2021real}
Xide Xia, Tianfan Xue, Wei-sheng Lai, Zheng Sun, Abby Chang, Brian Kulis, and
  Jiawen Chen.
\newblock Real-time localized photorealistic video style transfer.
\newblock In {\em Proceedings of the IEEE/CVF Winter Conference on Applications
  of Computer Vision}, pages 1089--1098, 2021.

\bibitem{xie2022neural}
Yiheng Xie, Towaki Takikawa, Shunsuke Saito, Or Litany, Shiqin Yan, Numair
  Khan, Federico Tombari, James Tompkin, Vincent Sitzmann, and Srinath Sridhar.
\newblock Neural fields in visual computing and beyond.
\newblock In {\em Computer Graphics Forum}, volume~41, pages 641--676. Wiley
  Online Library, 2022.

\bibitem{zakharov2019few}
Egor Zakharov, Aliaksandra Shysheya, Egor Burkov, and Victor Lempitsky.
\newblock Few-shot adversarial learning of realistic neural talking head
  models.
\newblock In {\em Proceedings of the IEEE/CVF international conference on
  computer vision}, pages 9459--9468, 2019.

\bibitem{zhou2019talking}
Hang Zhou, Yu Liu, Ziwei Liu, Ping Luo, and Xiaogang Wang.
\newblock Talking face generation by adversarially disentangled audio-visual
  representation.
\newblock In {\em Proceedings of the AAAI conference on artificial
  intelligence}, volume~33, pages 9299--9306, 2019.

\bibitem{zhou2022audio}
Jinxing Zhou, Jianyuan Wang, Jiayi Zhang, Weixuan Sun, Jing Zhang, Stan
  Birchfield, Dan Guo, Lingpeng Kong, Meng Wang, and Yiran Zhong.
\newblock Audio--visual segmentation.
\newblock In {\em European Conference on Computer Vision}, pages 386--403.
  Springer, 2022.

\bibitem{zhou2020makelttalk}
Yang Zhou, Xintong Han, Eli Shechtman, Jose Echevarria, Evangelos Kalogerakis,
  and Dingzeyu Li.
\newblock Makelttalk: speaker-aware talking-head animation.
\newblock {\em ACM Transactions on Graphics (TOG)}, 39(6):1--15, 2020.

\end{thebibliography}
}
\newpage

\section*{Supplementary Material}
\myparagraph{Overview.} In this supplementary material, we provide implementation details including the training setting and evaluation protocol~(Section~\hyperref[sec:a]{A}). Next, this supplementary material also provides an additional ablation study with a different set of hyperparameters to see their effects on the quality of audio-guided localized image stylization~(Section~\hyperref[sec:b]{B}). Lastly, in Section~\hyperref[sec:c]{C}, we provide more qualitative results with a variety of sound sources.

\section*{A. Implementation Details}
\subsection*{A.1. Audio-Visual Localizer}
\myparagraph{Training Details of Audio-Visual Encoder.} For training the audio encoder, we use the composited dataset as mentioned in the main paper. Each audio in the dataset is a 10-second log and sampled at frequencies of 44.1~kHz. Given a raw audio input, we preprocess audio inputs to produce Mel-spectrogram acoustic features. For each video clip, we randomly sample the middle frame and employ it as an input image~(30fps). We resize the middle frame to the image with resolution of 224$\times$224. We use Adam optimizer with the cosine cyclic learning rate scheduler~\cite{smith2017cyclical}. We set the learning rate to $10^{-4}$ with the momentum $0.9$ and weight decay $10^{-4}$. We use a batch size of 320. We adopt SpecAugment~\cite{park2019specaugment} for audio augmentation with the frequency mask ratio of $0.15$ and time masking ratio of $0.15$.

\myparagraph{Training Details of Audio-Visual Decoder.} We use the same composited dataset as the audio encoder is trained. We also preprocess audio inputs to produce Mel-spectrogram acoustic features. To generate a pseudo-segmentation mask, the corresponding text prompts are extracted from the short text descriptions~(e.g. people giggling). We adopt the following decoder architecture as CLIPSeg~\cite{luddecke2022image}, a zero-shot text-based segmentation model. Given the input image~352$\times$352, the pre-trained CLIP~\cite{radford2021learning} visual transformer projects activations of the internal layer to the 64 visual token embedding with patch size 16. The audio-visual decoder outputs the binary segmentation with a linear projection on the transformer's visual tokens at the last layer.

\myparagraph{Evaluation Details.} We compare our method to other sound source localization baselines. At the evaluation stage, the localization maps are resized to 224 $\times$ 224. For audio preprocessing, we extract the Mel-spectrograms and resize the audio input to 128$\times$512.

\subsection*{A.2. Audio-Guided INR Stylizer} 

\myparagraph{Training Details of INR.} 
We train our INR by using Multi-Layer Perceptron~(MLP) with 8 layers and 256 channels. Each MLP layer uses SIREN~\cite{sitzmann2020implicit} activation, except the output layer. We use $3$ channels for the output layer without an activation function. 
Following Tancik~\textit{et al.}~\cite{tancik2020fourier}, we use Gaussian mapping $\gamma$ for Fourier feature mapping function, which is computed as $\gamma(\mathbf{v})=[{\cos(2 \pi 10\mathbf{B} \mathbf{v}), \sin(2 \pi 10\mathbf{B} \mathbf{v})}]^\top$, where $\mathbf{v}$ denotes the pixel coordinate and each $\mathbf{B} \in \mathbb{R}^{m \times d}$ is sampled from $\mathcal{N}(0,\sigma^2)$ . We set $m$ as 256 and $d$ as 2, respectively. $\sigma$ is set to 1. We use Adam optimizer with a learning rate of $10^{-4}$. Stylizing one source image of size 512$\times$512 with our audio-guided INR stylizer takes one minute to train on a single GPU~(NVIDIA RTX 3090) for a total of 200 iterations. 

\begin{figure}[t!]
  \centering
  \includegraphics[width = \linewidth]{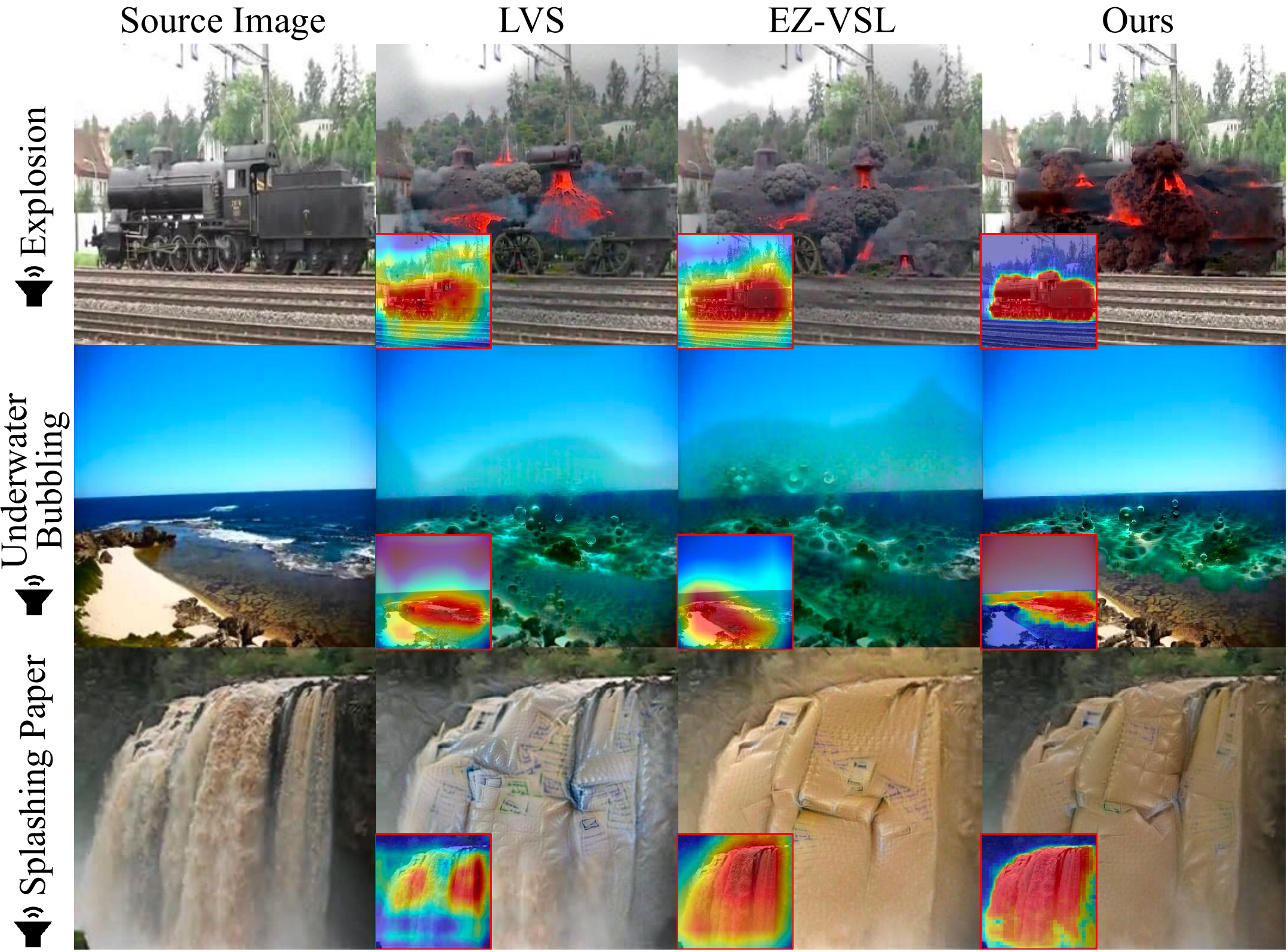}
  \caption{
   Stylization output comparision with existing sound source localization approaches~(LVS~\cite{chen2021localizing} and EZ-VSL~\cite{10.1007/978-3-031-19836-6_13}). The red box denotes the sound source localization map obtained by the method in each column with the given audio input. We impose~\textit{Train Sound, Wave Sound, Waterfall Sound} as localization conditions~(respectively, from top to bottom). We show that our sound source localization map is effective in stylizing the foreground more precisely than other baselines.
  }
  \vspace{-0.5em}
  \label{fig:sup_loc}
\end{figure}

\section*{B. Ablation Study}
\myparagraph{Effect of Foreground Regularization Loss.} To show the effect of foreground regularization loss, we conduct an ablation study. We use two different settings for the ablation study: training the INR without~$\mathcal{L}_\text{reg}$, without~$\mathcal{L}_\text{c}$. Fig.~\ref{fig:sup_reg} demonstrates that the stylized image can maintain the content of the source image by applying $\mathcal{L}_\text{c}, \mathcal{L}_\text{reg}$.  

\myparagraph{Effect of Audio-Visual Localizer.} We conduct an ablation study on applying different audio-visual localization maps to our localized style transfer~(see Fig.~\ref{fig:sup_loc}). We choose LVS~\cite{chen2021localizing} and EZ-VSL~\cite{10.1007/978-3-031-19836-6_13} as the baselines. Then, we show that our localization map helps create the locally stylized images. In particular, our method prevents rough stylization at the boundary of the object.

\myparagraph{Effect of INR Optimization for Local Area.} We compare our method to a na\"ive approach that attaches globally stylized images to the source images. As Fig.~\ref{fig:figure_sup_cutpaste} shows, we observe that optimizing the INR over the localized area helps the semantics of the audio to appear in the foreground. For example, given the pond image, and the target sound~\textit{``Grass Scratching''}, we stylize an image over the entire area~(see Fig.~\ref{fig:figure_sup_cutpaste}~(a)) and then multiply the localization map for the sound of~\textit{``Water''}~(see Fig.~\ref{fig:figure_sup_cutpaste}~(b)). As shown in Fig.~\ref{fig:figure_sup_cutpaste}~(c), the foreground of the stylized image is more highly associated with this audio semantic.

\begin{figure}[t!]
  \centering
  \includegraphics[width = \linewidth]{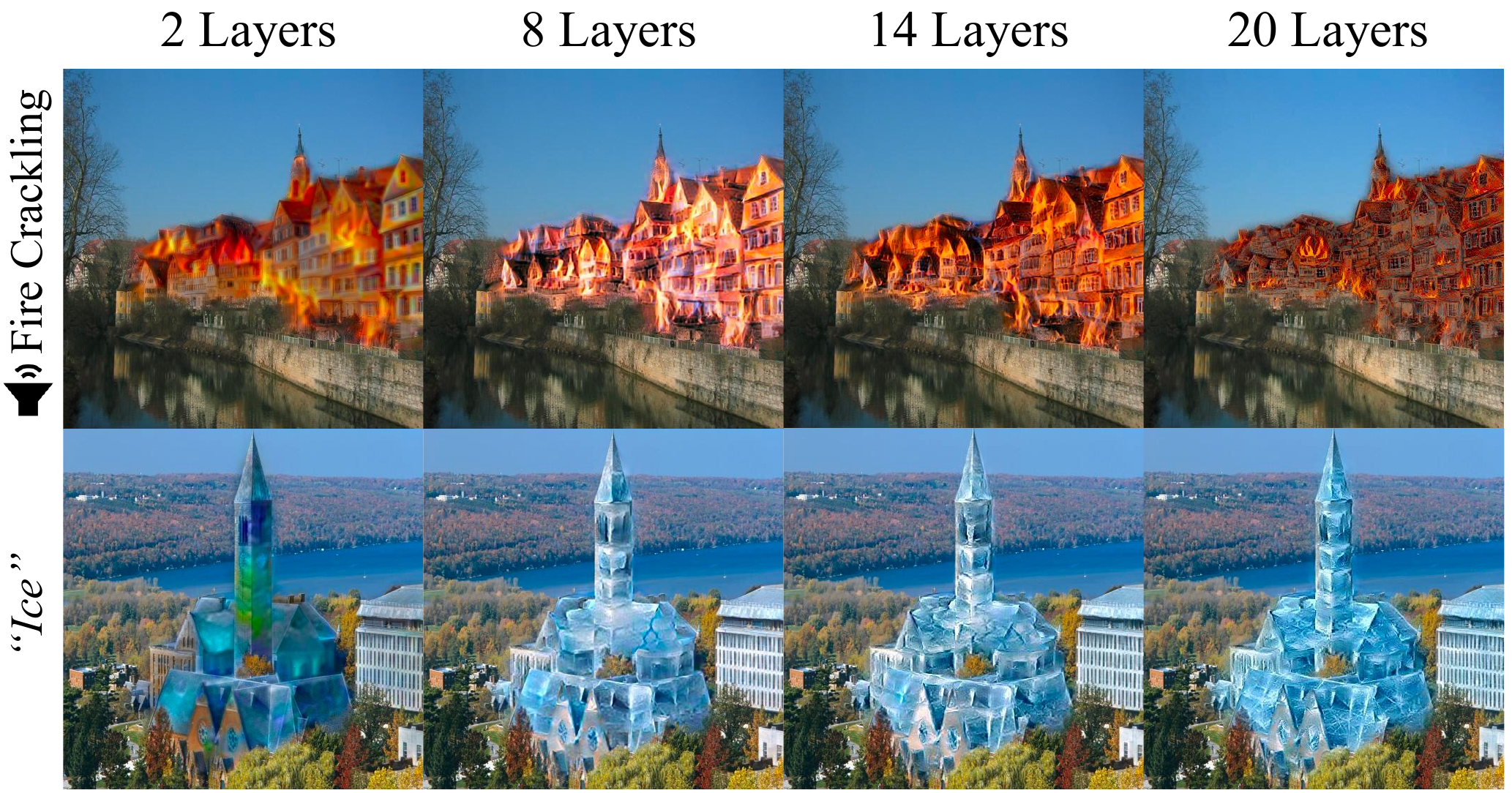}
  \caption{
   Localized image stylization results according to the number of SIREN~\cite{sitzmann2020implicit} layers. 
  }
  \vspace{-0.5em}
  \label{fig:sup_figure_ablation_mlp}
\end{figure}

\section*{C. Qualitative and Quantitative Results} 

\myparagraph{The Number of SIREN Layers.} As mentioned in the paper, we analyze the effect according to the number of fully-connected SIREN~\cite{sitzmann2020implicit} layers. As Fig.~\ref{fig:sup_figure_ablation_mlp} shows, we compare the stylization results by increasing the number of SIREN layers to 2, 8, 14, and 20. When the number of layers is two, INR cannot represent the style in the foreground. However, when the number of layers is increased to 8, the style in the foreground becomes more realistic. 
As the number of layers increases, more iterations are required for convergence, resulting in more time consumption. Specifically, the required time to optimize the INR is 37, 48, 61, and 72 seconds, as the number of layers is 2, 8, 14, and 20, respectively.

% AMT Capture. Metric. 

\myparagraph{Stylizing High-Resolution Image.} As  Fig.~\ref{fig:sup_highresolution2} shows, we also perform localized image stylization with high-resolution input images. We compare the stylization results of CLIPstyler~\cite{kwon2022clipstyler} and our method. We observe that our stylized image is more realistic than the CLIPstyler. There are two reasons: multi-scale patching and INR. While CLIPstyler uses too small patches for the given high-resolution image, our method adjusts the patch size for stylization. Moreover, our approach can capture high-frequency details using INR, which can stylize the source image even in high resolution.

\myparagraph{Additional Qualitative and Quantitative Examples.} Fig~\ref{fig:sup_interactive_showcase} shows interactive stylization results with different types of condition modality using multi-modal joint embedding space. We provide the stylization results for four combinations of audio and text as follows:

\begin{itemize}
    \item Audio-localized \& audio-guided image stylization. \vspace{-.5em}
    \item Audio-localized \& text-guided image stylization. \vspace{-.5em}
    \item Text-localized \& audio-guided image stylization. \vspace{-.5em}
    \item Text-localized \& text-guided image stylization. 
\end{itemize}
Furthermore, we compare our method with text-driven localized image stylization, Text2LIVE~\cite{bar2022text2live}. Since audio and text and image share the CLIP embedding space, our method can also stylize the image with the user-provided text descriptions. First, we create a probability mask with CLIPSeg, a zero-shot text-based segmentation. Then, we perform localized image stylization with our INR-based optimization. 
As shown in Fig~\ref{fig:sup_text2live}, we observe that Text2LIVE shows distorted structures on the outside of the target area, whereas our method produces a natural image without disturbing background areas.

\begin{figure}[t!]
  \centering
  \includegraphics[width = \linewidth]{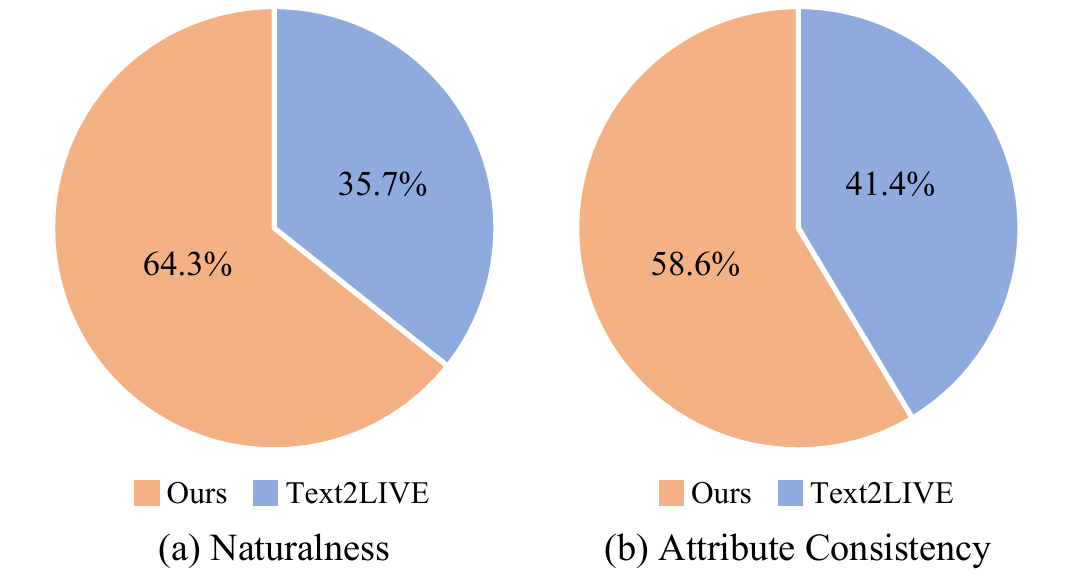}
  \caption{
  User study results comparing LISA vs. Text2LIVE~\cite{bar2022text2live}. We use the same questionnaire as described in the main paper: (a) Naturalness and (b) Attribute Consistency.
  }
  \vspace{-0.5em}
  \label{fig:sup_userstudy}
\end{figure}

\begin{figure*}[t!]
  \centering
  \includegraphics[width = \linewidth]{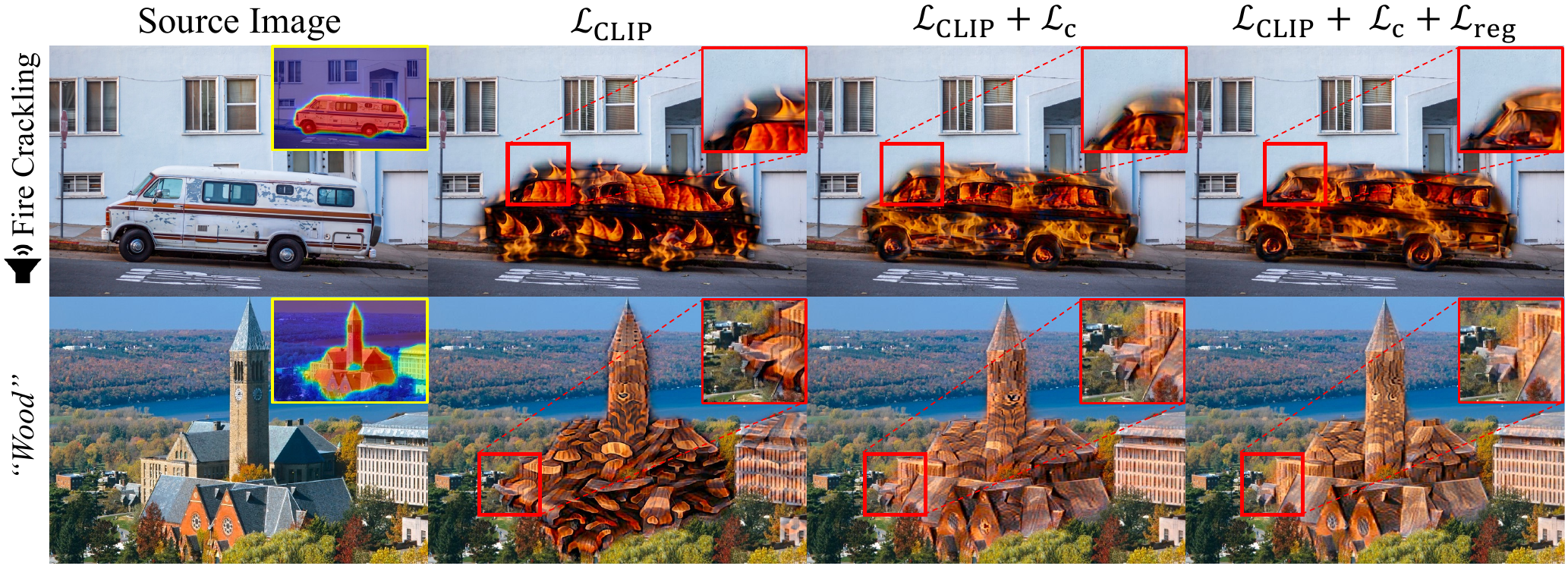}
  \caption{
   Effect of our Foreground Regularization loss. $\mathcal{L}_\text{c}$, $\mathcal{L}_\text{reg}$ preserve the visual properties of the source image. The yellow box denotes the localization map conditioned by \textit{Car Sound} and \textit{``church"} respectively from top to bottom. The red box denotes a magnified image of the border area in each stylized image.
%   fully supervised approach.
  }
%   \vspace{-0.5em}
  \label{fig:sup_reg}
\end{figure*}

\begin{figure*}[t!]
  \centering
  \includegraphics[width = \linewidth]{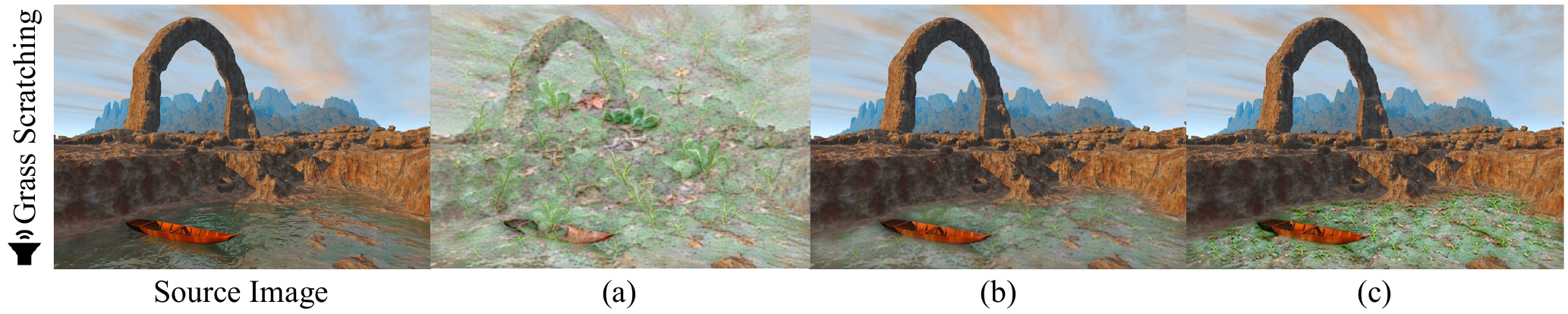}
  \caption{
   Effect of INR optimization for the local area. 
   (a), (b), and (c) refers to sound-guided global stylization, an image pasted into the source image by cutting the mask area from (a), and our local area optimization approach, respectively. While (b) fails to stylize the local area, our method (c) stylizes the semantics of the audio to be well-matched to the local area.
   }
%   \vspace{-0.5em}
  \label{fig:figure_sup_cutpaste}
\end{figure*}

\begin{figure*}[t!]
  \centering
  \includegraphics[width = 18cm]{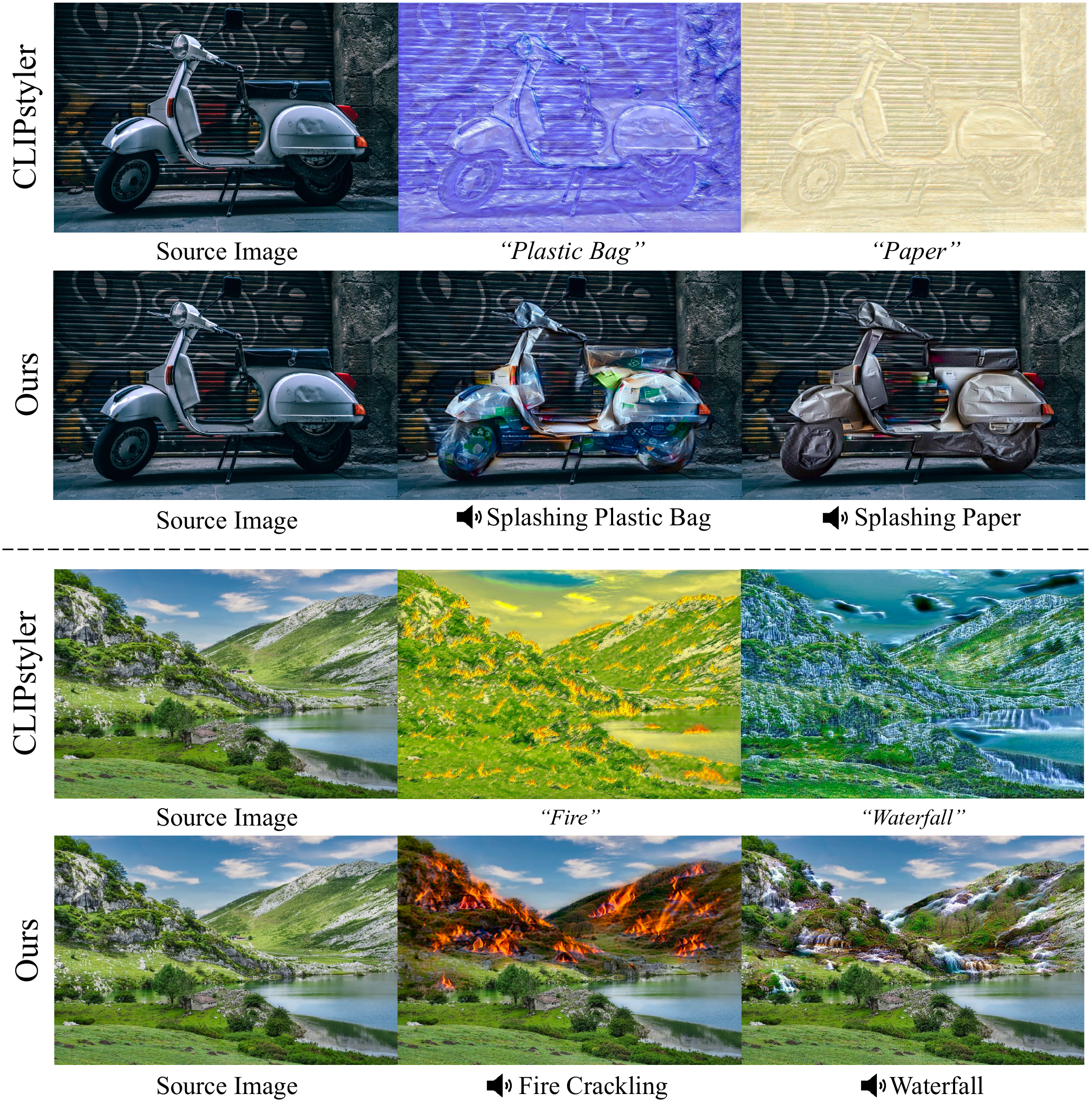}
  \caption{
  Comparison of stylizing high-resolution image $(1536 \times 1024)$ using CLIPstyler~\cite{kwon2022clipstyler} and ours. Unlike CLIPstyler, our method can generate a realistic style even for a high-resolution image.
%   fully supervised approach.
  }
%   \vspace{-0.5em}
  \label{fig:sup_highresolution2}
\end{figure*}

\begin{figure*}[t]
    \begin{center}
        \includegraphics[width=\textwidth]{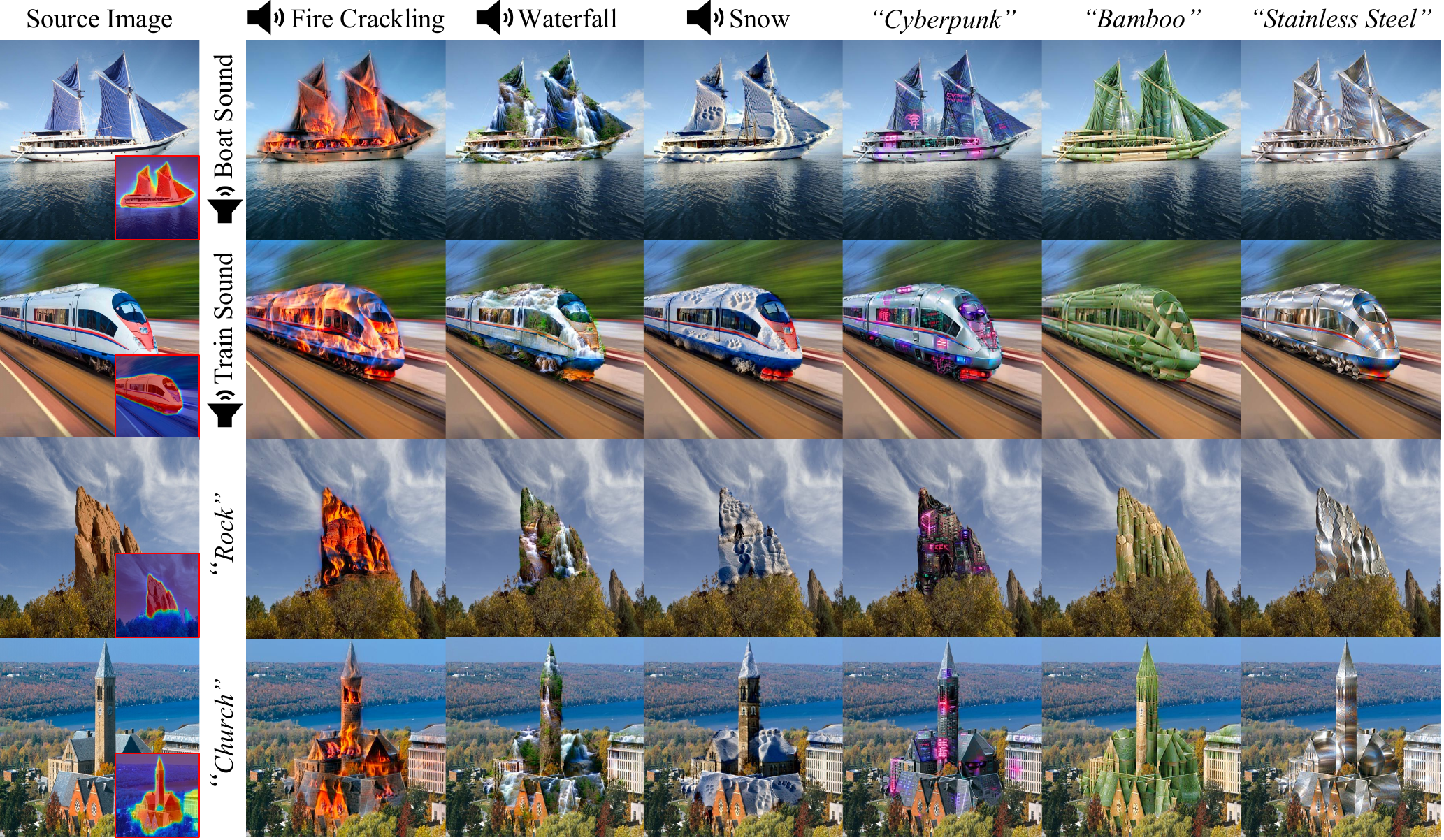}
    \end{center}
    \caption{Interactive stylization results with different types of condition modality. The first to second rows use audio as a localization condition, and from the third to last rows localization is conditioned by text prompt. For stylization conditions, from second to fourth columns are conditioned by audio, and from fifth to the final column is conditioned by text prompts. The red box at each source image indicates a predicted sound source localization map, respectively. }
    \label{fig:sup_interactive_showcase}
\end{figure*}

\begin{figure*}[t!]
  \centering
  \includegraphics[width=\linewidth]{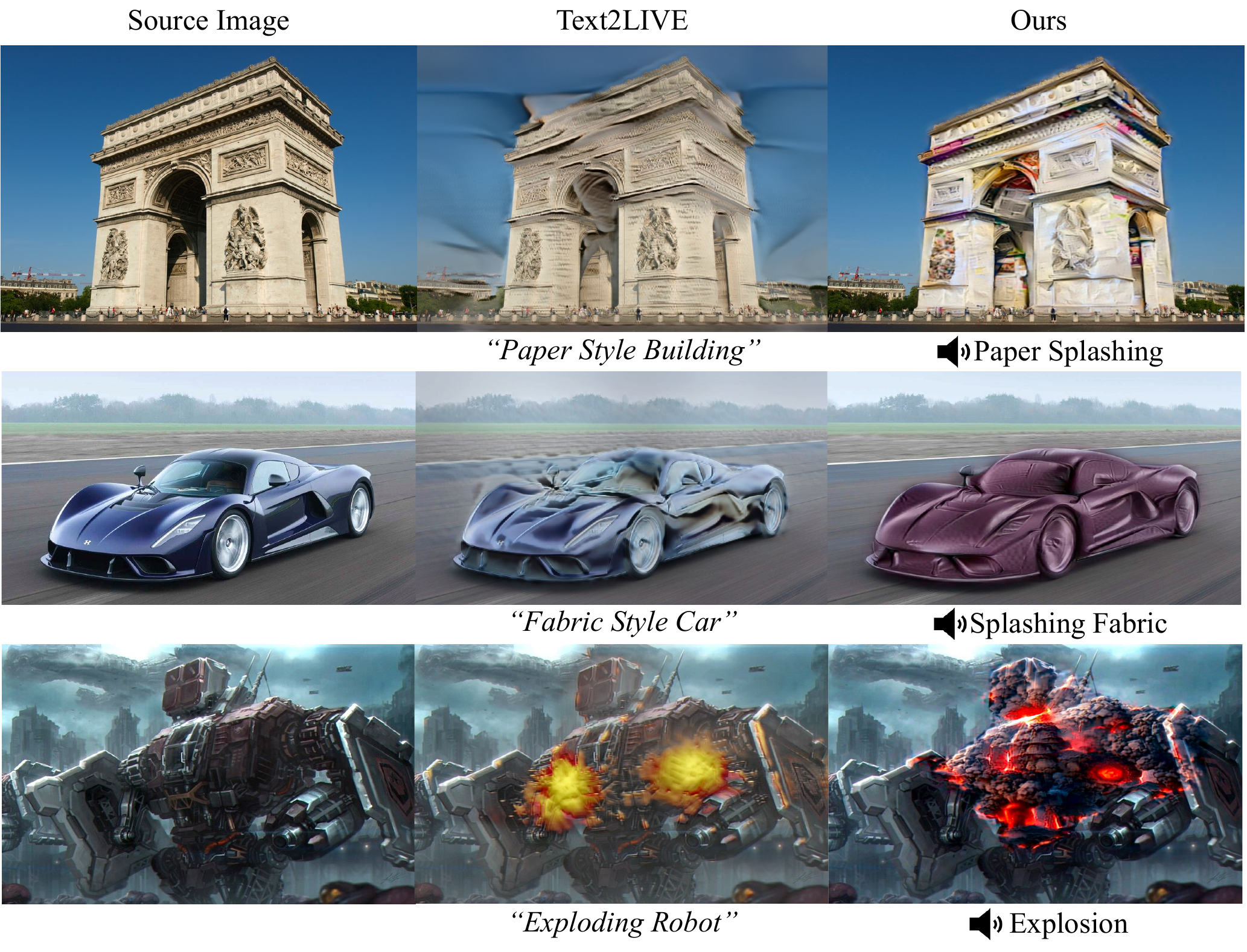}
  \caption{
   Additional comparison between Text2LIVE~\cite{bar2022text2live} and Ours. 
   Text2LIVE shows distorted structures on the outside of the target area. However, our approach stylizes the foreground of the source image per-pixel with maintaining the background.
%   fully supervised approach.
  }
  \label{fig:sup_text2live}
\end{figure*}

\myparagraph{Additional User Study.} We conduct an additional user study to compare our method with Text2LIVE~\cite{bar2022text2live}. As mentioned in the paper, we assemble 100 participants from Amazon Mechanical Turk~(AMT). For a fair comparison, we perform text-localized image stylization. Fig~\ref{fig:sup_userstudy} shows that our method is competitive to text-localized image stylization. 

\myparagraph{User Study Details.} Our user study contains two parts for the evaluation of naturalness and attribute consistency. Each section consists of 20 random
binary choice questions that compare our method to one of the baselines~(AVStyle~\cite{li2021learning} and Lee~\textit{et al}~\cite{lee2022sound} and CBStyling~\cite{kurzman2019class}). We emphasize that our user study is anonymous and does not collect any personally identifiable data. % Additionally, we compare our method with Text2LIVE. 

\end{document}